\documentclass[final,3p,times,twocolumn,authoryear]{elsarticle}
\usepackage{amsmath,amsfonts,amssymb}
\usepackage{stfloats}
\usepackage{algpseudocode}
\usepackage{algorithm}
\usepackage{array}
\usepackage{textcomp}
\usepackage{url}
\usepackage{verbatim}
\usepackage{graphicx}
\usepackage{booktabs}
\usepackage{multirow}
\usepackage[table,xcdraw,dvipsnames]{xcolor}
\usepackage{colortbl}
\usepackage{caption}
\usepackage{subcaption}
\usepackage{tikz}

\usetikzlibrary{shapes.geometric, arrows.meta}

\newcommand{\RefTarget}{
    \tikz[baseline=-0.6ex]{
        \node[star, star points=5, star point ratio=2.25, fill=red, draw=black, line width=0.2pt, inner sep=1.2pt] {};
    }
}
\newcommand{\RefCurrentResidual}{
    \tikz[baseline=-0.6ex]{
        \draw[orange, densely dotted, line width=1.2pt] (0,0) -- (0.3,0);
    }
}
\newcommand{\RefPreviousResidual}{
    \tikz[baseline=-0.6ex]{
        \draw[red, dashed, line width=1.2pt] (0,0) -- (0.3,0);
    }
}
\newcommand{\RefStep}{
    \tikz[baseline=-0.6ex]{
        \draw[blue, -{Stealth[length=1.5mm]}, line width=1.2pt] (0,0) -- (0.3,0);
    }
}
\newcommand{\RefSubspace}{
    \tikz[baseline=-0.6ex]{
        \fill[magenta, opacity=0.3] (0,-0.1) rectangle (0.3,0.2);
        \draw[purple, line width=0.5pt] (0,-0.1) rectangle (0.3,0.2);
    }
}
\newcommand{\RefPosition}{
    \tikz[baseline=-0.6ex]{
        \fill[black] (0,0) circle (1.5pt);
    }
}
\newcommand{\RefHistory}{
    \tikz[baseline=-0.6ex]{
        \draw[gray, line width=1pt] (0,0) -- (0.3,0);
        \fill[gray] (0.3,0) circle (1pt);
    }
}     
\newcommand{\RefTrajectory}{
    \tikz[baseline=-0.6ex]{
        \draw[gray, opacity=0.8, line width=1pt] (0,0) -- (0.5,0);
    }
}
\definecolor{mpl_green}{HTML}{008000}
\newcommand{\RefStart}{%
    \tikz[baseline=-0.6ex]{
        \fill[mpl_green] (0,0) circle (2pt);
    }
}
\newcommand{\RefMean}{
    \tikz[baseline=-0.6ex]{
        \draw[orange, line width=1.5pt] (-0.1,-0.1) -- (0.1,0.1);
        \draw[orange, line width=1.5pt] (-0.1,0.1) -- (0.1,-0.1);
    }
}
\newcommand{\mylegend}[2]{
  \raisebox{0.2em}{\colorbox{#1}{\rule{0pt}{2pt}\rule{2pt}{0pt}}}\hspace{0.5em}#2
}
\newcommand{\RefErrorBar}{
    \tikz[baseline=-0.6ex, line width=0.3pt, black, line cap=round]{
        \draw (0, -0.6ex) -- (0, 1ex); 
        \draw (-0.3ex, -0.6ex) -- (0.3ex, -0.6ex); 
        \draw (-0.3ex, 1ex) -- (0.3ex, 1ex); 
    }
}

\definecolor{RSB}{HTML}{B07AA1}
\definecolor{RF}{HTML}{4E79A7}
\definecolor{XGB}{HTML}{E15759}
\definecolor{LGBM}{HTML}{76B7B2}
\definecolor{CatBoost}{HTML}{59A14F}

\definecolor{wil_bon_3}{HTML}{4A1856}
\definecolor{wil_bon_2}{HTML}{7F286D}
\definecolor{wil_bon_1}{HTML}{B3416A}
\definecolor{wil_bon_0}{HTML}{DF6961}
\definecolor{wil_bon_no_sign}{HTML}{E8E8E8}

\begin{document}

\begin{frontmatter}

\journal{Expert Systems with Applications}

\title{Random Symbolic Boosting: A Geometric Approach to Symbolic Regression}


\author[a]{José R. Ortega\corref{cor1}}
\cortext[cor1]{Corresponding author}
\ead{jose.r.ortega94@gmail.com}
\author[b]{José Manuel Muñoz Contreras}
\author[a,c]{Leonardo Trujillo}

\affiliation[a]{organization={Tecnológico Nacional de México/Instituto Tecnológico de Tijuana},
            addressline={Blvd. Limón Padilla (Industrial) y Ave. ITR Tijuana s/n, Mesa de Otay},
            postcode={22500},
            city={Tijuana},
            state={Baja California},
            country={México}}

\affiliation[b]{organization={Universidad Autónoma de Baja California, Unidad Valle de las Palmas},
            addressline={Calzada Universidad 14418, Parque Industrial Internacional Tijuana},
            postcode={22427},
            city={Tijuana},
            state={Baja California},
            country={México}}

\affiliation[c]{organization={LASIGE, Department of Informatics, Faculty of Sciences, University of Lisbon},
            addressline={Campo Grande},
            postcode={1749-016},
            city={Lisbon},
            country={Portugal}}

\begin{highlights}
    \item RSB builds explicit symbolic models by boosting random syntax trees.
    \item Training error decreases monotonically for OLS in regression and classification.
    \item An angle-based early stopping criterion is independent of the problem scale.
    \item Ranks 1st on real-world SRBench problems; Pareto-optimal accuracy vs. training time.
    \item RSB is competitive with leading black-box models on 200 TALENT problems.
\end{highlights}

\begin{abstract}

Symbolic models offer interpretability, but their construction through generational evolution can be computationally expensive. This article presents Random Symbolic Boosting (RSB), a method that constructs explicit additive symbolic models via boosting over populations of randomly generated syntax trees, eliminating the need for generational evolution. The idea combines two observations: high-dimensional spaces can be spanned by enough linearly independent vectors, and the semantics of syntax trees can be directly manipulated as such vectors. It is shown that, for Ordinary Least Squares, the boosting process guarantees a monotonic reduction of the training error for both regression and classification. Furthermore, an adaptive geometric early stopping mechanism is introduced. It is based on the angle between the prediction and target vectors, is independent of the problem's scale, and requires no adjustments across tasks. Evaluated on a standard benchmark for symbolic regression, the method is located at the Pareto-optimal point between predictive accuracy and training time, ranking first on the real-world datasets. Additionally, it was tested on a large tabular classification benchmark with 200 problems, achieving competitive results.
\end{abstract}

\begin{keyword}
Symbolic Regression \sep Gradient Boosting \sep Semantic Genetic Programming \sep Early Stopping

\end{keyword}

\end{frontmatter}

\section{Introduction}\label{intro}

There are two main approaches in data modeling through supervised learning \citep{rudin2019stop}. The first, and most popular, is the use of black-box models, such as neural networks or decision-tree ensemble methods, which are fast and accurate, but their internal logic is not directly interpretable. The second approach seeks interpretability through symbolic models, such as Genetic Programming (GP) \citep{koza1994genetic} or its semantic variant, Geometric Semantic Genetic Programming (GSGP) \citep{moraglio2012geometric}, but the generational, population-based process of these methods is computationally expensive. In black-box models, optimization is usually restricted to tuning model parameters or finding decision thresholds, which is more efficient than evolving symbolic structures. However, this can lead to local optima that are only corrected by changing the model \citep{elsken2019neural}. 

To address these limitations, the Random Symbolic Boosting (RSB) method is proposed. Although the modification or elimination of standard generational evolution has been explored by previous approaches, such as Fast Function Extraction (FFX) \citep{mcconaghy2011ffx} and Kaizen Programming (KP) \citep{demelo2014kaizen}, RSB achieves state-of-the-art performance using a boosting scheme \citep{friedman2001greedy} that fits linear combinations of the semantics of stochastically generated syntax trees. This reduces the computational cost by replacing GP's evolutionary search with an optimization over the residuals. The result is an explicit symbolic model that achieves competitive accuracy on regression and classification tasks.

Furthermore, an early stopping mechanism based on the geometry of the semantic subspace is introduced. Instead of evaluating an unbounded metric to decide when to stop, the angle formed between the target and the prediction is calculated, which, in addition to being bounded, is applicable to supervised learning tasks.

The remainder of the article is organized as follows: Section \ref{background} reviews related work. Section \ref{RSB} describes the RSB method. Sections \ref{experimental} and \ref{benchmarking} present the experimental analysis and the evaluation of its predictive performance, respectively. Finally, Sections \ref{discussions} and \ref{conclusions} discuss the results and conclusions.

\section{Related Work}\label{background}

GP is the classical method for generating symbolic models through evolution \citep{koza1994genetic}. However, it often suffers from convergence problems. GSGP addresses this by working directly on the semantics of the solutions, inducing a unimodal error surface \citep{moraglio2012geometric}. GSGP can also be seen as building each offspring as a combination of its parents’ semantics \citep{moraglio2012geometric}, and \citet{dick2024gsgp_ensemble} interprets it as ensemble learning, linking semantic mutation and crossover to boosting and stacking. Semantic backpropagation GP (SBP-GP) \citep{virgolin2019sbp} also operates on semantics, but defines its variation operators by inverting program execution to obtain the desired semantics at selected nodes \citep{pawlak2015semantic}. Hybrid methods such as Multiple Regression Genetic Programming (MRGP) \citep{arnaldo2014multiple} and MAP-Elites Genetic Programming (MEGP) \citep{zhang2023map} integrate linear regression and diversity maps, respectively, with the goals of improving generalization capability and enhancing search space exploration. Along the lines of MRGP, Parallel Multidimensional Genetic Programming with Multidimensional Populations for Symbolic Regression (M5GP) \citep{cardenas2024m5gp} focuses on evolving multidimensional feature transformations through a stack-based representation with GPU parallel processing, where the new semantic dimensions are subsequently evaluated by a multiple linear regression model to produce the final prediction. Nevertheless, all these variants still rely on evolutionary search to obtain their solutions.

Beyond evolving a single solution, empirical evidence suggests that the aggregation of multiple simple models frequently outperforms algorithms that search for a single complex solution \citep{NEURIPS2022_0378c769}. In symbolic regression, algorithms such as FFX \citep{mcconaghy2011ffx}, which is deterministic, and KP \citep{demelo2014kaizen}, which uses a collaborative evolutionary approach, build models through the sum of basis functions. Specifically, FFX performs a systematic and exhaustive expansion of basis functions that it subsequently prunes via deterministic minimization with regularization (Pathwise Regularized Learning). Similarly, Evolutionary Feature Synthesis (EFS) \citep{arnaldo2015efs} evolves nonlinear features bottom-up and selects a sparse subset via pathwise regularized regression, yielding a readable linear combination of composed features without a symbolic tree representation. By contrast, KP initializes a team of experts and applies statistical significance tests, based on $p$-values and coefficient magnitude thresholds, to filter terms and determine the contribution of each. This follows the same logic as decision tree-based methods like Random Forest or XGBoost \citep{NEURIPS2022_0378c769}, where the error is reduced through the aggregation of multiple models. Symbolic Regression-based Random Forest (SR-Forest) \citep{zhang2024srforest} attempts to combine both approaches by constructing a forest of hybrid models; each consists of a Ridge regression that linearly combines a set of syntax trees, and a random decision tree applied to the residuals of this regression. The algorithm applies evolutionary operators to the sets of syntax trees, using a bagging strategy to iteratively retain and combine complementary models.

The integration of semantic and boosting techniques has also been explored by Symbolic-regression boosting (SyRBo) \citep{sipper2021syrbo}; this method modifies standard gradient boosting by replacing traditional weak estimators with GP-based symbolic regressors. Similarly, Semantic Boosting Regression (SBR) \citep{leite2025sbr} sequentially aggregates Memetic Semantic Genetic Programming (MSGP) \citep{ffrancon2015memetic} models, using semantic backpropagation for local refinement.

A limitation of most methods is the computational cost inherent to the structural search and evaluation. For example, \citet{lacava2021contemporary} reports average times close to $10^5$ seconds to generate a model with MRGP, while in the case of SR-Forest, its authors report that the process requires an average of $10^4$ seconds. However, Reservoir Computing (RC) \citep{lukosevicius2009reservoir} demonstrates that it is not necessary to optimize the internal structure of the model; it is sufficient to project the data into a high-dimensional random space and only fit the final output.

RSB integrates these principles; it generates a population of random syntax trees that act as projections into a semantic space. However, instead of using a static reservoir, RSB builds the model step by step through a linear combination optimized with boosting and deterministically selects the sets of trees that best reduce the residual error, following the additive logic of XGBoost.

\section{Proposed Method}\label{RSB}

\subsection{Theoretical Framework}

Consider a training dataset $(\mathbf{X}_{tr}, \mathbf{y}_{tr})$, where $\mathbf{X}_{tr} = [\mathbf{x}_1, \dots, \mathbf{x}_n]^\top \in \mathbb{R}^{n \times d}$ is the input matrix containing $n$ feature vectors $\mathbf{x}_i \in \mathbb{R}^d$, and $\mathbf{y}_{tr} = [y_1, \dots, y_n]^\top \in \mathbb{R}^n$ denotes the target vector, with $y_i \in \mathbb{R}$ for regression and $y_i \in \{0,1\}$ for binary classification. The RSB algorithm constructs an additive model $F_K : \mathbb{R}^d \to \mathbb{R}$ through the sequential aggregation of $K$ weak estimators
\begin{equation}
    F_K(\mathbf{x}) = \sum_{k=1}^K \eta\, u^{(k)}(\mathbf{x}),
    \label{eq:additive_model}
\end{equation}
\noindent where $\eta$ represents the learning rate (shrinkage), $u^{(k)} : \mathbb{R}^d \to \mathbb{R}$ denotes the update function of level $k$, and $\mathbf{x} \in \mathbb{R}^d$ is an arbitrary feature vector. The procedure for regression and binary classification problems is detailed in Algorithm \ref{alg:RSB_general}.

\begin{algorithm}[h]
\caption{Random Symbolic Boosting (RSB)}
\label{alg:RSB_general}
\footnotesize
\begin{algorithmic}[1]
    \Require $\mathbf{X}_{tr}$, $\mathbf{y}_{tr}$, $\mathbf{X}_{val}$, $\mathbf{y}_{val}$, task, iterations $K$, population $M$, learning rate $\eta$, Elastic Net parameters ($\lambda, \rho$), Early Stopping parameters ($W$, $\gamma$, $P_{c}$, $P_{d}$, $L$, $\beta_{tol}$), function set $\mathcal{F}$, terminal set $\mathcal{Z}$, max depth $d_{max}$, terminal probability $p_t$, constant probability $p_c$.
    \Ensure set of predictors $\mathcal{P}$.
    \State $\mathcal{P} \gets \emptyset$, $\mathbf{F}_{tr} \gets \mathbf{0}$, $\mathbf{F}_{val} \gets \mathbf{0}$
    \State $E_{best} \gets \infty$, $\mathcal{P}_{best} \gets \emptyset$, $k_{best} \gets 0$
    
    \For{$k = 1$ \textbf{to} $K$}
        \State $\mathbf{r} \gets \textsc{ComputeResidual}(\mathbf{y}_{tr}, \mathbf{F}_{tr}, \text{task})$ \Comment{Eq. \ref{eq:residuals}}
        
        \State $\mathcal{T}^{(k)} \gets \{T^{(k)}_1, \dots, T^{(k)}_M\} \sim \textsc{GenTree}(\mathcal{F}, \mathcal{Z}, d_{max}, p_t, p_c)$
        \State $\mathbf{S}_{tr} \gets [T^{(k)}_1(\mathbf{X}_{tr}), \dots, T^{(k)}_M(\mathbf{X}_{tr})]$
        
        \State $(\tilde{\mathbf{S}}_{tr},\, \boldsymbol{\nu}_k,\, \boldsymbol{\delta}_k) \gets \textsc{Transform}(\mathbf{S}_{tr})$ \Comment{Eq. \ref{eq:transform}}
        
        \State $\boldsymbol{\alpha}^*_k \gets \textsc{SolveElasticNet}(\tilde{\mathbf{S}}_{tr}, \mathbf{r}, \lambda, \rho)$ \Comment{Eq. \ref{eq:elastic_net}}
        
        \State $\mathbf{u}_{tr}^{(k)} \gets \tilde{\mathbf{S}}_{tr}\boldsymbol{\alpha}^*_k$
        \State $\mathbf{F}_{tr} \gets \mathbf{F}_{tr} + \eta \mathbf{u}_{tr}^{(k)}$ \Comment{Eq. \ref{eq:model_update}}
        
        \State $\mathbf{S}_{val} \gets [T^{(k)}_1(\mathbf{X}_{val}), \dots, T^{(k)}_M(\mathbf{X}_{val})]$
        \State $\tilde{\mathbf{S}}_{val} \gets (\mathbf{S}_{val} - \mathbf{1}\boldsymbol{\nu}_k^\top)\operatorname{diag}(\boldsymbol{\delta}_k)^{-1}$
        \State $\mathbf{u}_{val}^{(k)} \gets \tilde{\mathbf{S}}_{val}\boldsymbol{\alpha}^*_k$
        \State $\mathbf{F}_{val} \gets \mathbf{F}_{val} + \eta \mathbf{u}_{val}^{(k)}$ \Comment{Eq. \ref{eq:model_update}}
        
        \State $\mathcal{P} \gets \mathcal{P} \cup \{(\mathcal{T}^{(k)}, \boldsymbol{\alpha}^*_k, \boldsymbol{\nu}_k, \boldsymbol{\delta}_k)\}$
        
        \State $E_{val} \gets \textsc{ComputeError}(\mathbf{y}_{val}, \mathbf{F}_{val}, \text{task})$ \Comment{RMSE or Log-Loss}
        \If{$E_{val} < E_{best}$}
            \State $E_{best} \gets E_{val}$;\quad $\mathcal{P}_{best} \gets \mathcal{P}$;\quad $k_{best} \gets k$
        \EndIf

        \State $\theta_k \gets \textsc{CalculateAngle}(\mathbf{F}_{val}, \mathbf{y}_{val}, \text{task})$ \Comment{Eq. \ref{eq:theta}}
        
        \State $\text{status} \gets \textsc{CheckEarlyStopping}(\theta_k, W, \gamma, P_{c}, P_{d}, L, \beta_{tol})$
        \If{$\text{status} == \text{CONVERGENCE}$}
            \State \textbf{break} \Comment{Keeps current $\mathcal{P}$ up to $k$}
        \ElsIf{$\text{status} == \text{DEGRADATION}$}
            \State $\mathcal{P} \gets \mathcal{P}_{best}$ \Comment{Rollback to best level $k_{best}$}
            \State \textbf{break}
        \EndIf
    \EndFor
    \State \Return $\mathcal{P}$
\end{algorithmic}
\end{algorithm}

The process is initialized with $F_0(\mathbf{x}) = 0\ \forall\, \mathbf{x}$. At each iteration $k$, for each instance $i$, the residual $r_{k,i} \in \mathbb{R}$ is calculated, which is defined by the nature of the task
\begin{equation}
    r_{k,i} = 
    \begin{cases}
        y_i - F_{k-1}(\mathbf{x}_i) & \textit{Regression} \\
        y_i - \sigma(F_{k-1}(\mathbf{x}_i)) & \textit{Binary Classification}
    \end{cases}.
    \label{eq:residuals}
\end{equation}
\noindent where $\sigma(\cdot)$ is the sigmoid function. Subsequently, a random population of $M$ syntax trees $\{T_m^{(k)} : \mathbb{R}^d \rightarrow \mathbb{R}\}_{m=1}^M$ with a maximum depth of $d_{max}$ is generated. Their evaluation on a vector $\mathbf{x}$ produces the semantic vector $\mathbf{s}^{(k)} : \mathbb{R}^d \rightarrow \mathbb{R}^M$
\begin{equation}
    \mathbf{s}^{(k)}(\mathbf{x}) = \bigl[ T_1^{(k)}(\mathbf{x}), \dots, T_M^{(k)}(\mathbf{x}) \bigr]^\top \in \mathbb{R}^{M},
    \label{eq:semantic_vector}
\end{equation}
and the concatenation of the $n$ semantic vectors forms the semantic matrix
\begin{equation}
    \mathbf{S}^{(k)} = \bigl[ \mathbf{s}^{(k)}(\mathbf{x}_1), \mathbf{s}^{(k)}(\mathbf{x}_2), \ldots, \mathbf{s}^{(k)}(\mathbf{x}_n) \bigr]^\top \in \mathbb{R}^{n\times M}.
    \label{eq:semantic_matrix}
\end{equation}

To handle the diversity of dynamic ranges in the generated semantics, the matrix $\mathbf{S}^{(k)}$ undergoes a column-wise linear transformation, denoted as $\tilde{\mathbf{S}}^{(k)}$. This operation allows the regularization to treat all trees equally and is defined as
\begin{equation}
    \tilde{\mathbf{S}}^{(k)} = \bigl(\mathbf{S}^{(k)} - \mathbf{1}_n\boldsymbol{\nu}_k^\top\bigr)\operatorname{diag}(\boldsymbol{\delta}_k)^{-1},
    \label{eq:transform}
\end{equation}
\noindent where $\boldsymbol{\nu}_k$ and $\boldsymbol{\delta}_k \in \mathbb{R}^M$ represent the translation and scaling parameters of each column (for example, the mean and standard deviation (SD) in the case of standardization, or the minimum and the range in Min-Max scaling), and $\mathbf{1}_n \in \mathbb{R}^n$ is the vector of ones. It is assumed that $[ \boldsymbol{\delta}_k ]_m \neq 0\ \forall m$; otherwise, it is replaced by $\big[ \boldsymbol{\delta}_k \big]_m = 1$ for the corresponding tree. If no transformation is required, $\boldsymbol{\nu}_k$ is the null vector and $\boldsymbol{\delta}_k = \mathbf{1}_M$.

Subsequently, Elastic Net \citep{zou2005regularization} is employed—an approach also utilized by deterministic algorithms such as FFX—to obtain the optimal coefficient vector $\boldsymbol{\alpha}_k^*$. Elastic Net generalizes the configurations of Ordinary Least Squares (OLS) with $\lambda = 0$, Lasso or $L_1$ norm ($\lambda > 0, \rho = 1$), and Ridge or $L_2$ norm ($\lambda > 0, \rho = 0$) as special cases; its general form is defined as
\begin{equation}
    \begin{aligned}
        \boldsymbol{\alpha}^*_k =  \arg\min_{\boldsymbol{\alpha} \in \mathbb{R}^M} \Biggl( & \frac{1}{2n} \| \tilde{\mathbf{S}}^{(k)}\boldsymbol{\alpha} - \mathbf{r}_{k} \|_2^2 \\
        & + \lambda \rho \|\boldsymbol{\alpha}\|_1 + \frac{\lambda(1-\rho)}{2} \|\boldsymbol{\alpha}\|_2^2 \Biggr),
    \end{aligned}
    \label{eq:elastic_net}
\end{equation}
\noindent where $\mathbf{r}_k=\big[ r_{k,1}, r_{k,2}, \ldots, r_{k,n} \big]^\top \in \mathbb{R}^n$, $\lambda \geq 0$ controls the penalty intensity, and $\rho \in [0, 1]$ establishes the balance between the $L_1$ and $L_2$ norms. The candidate update for the boosting level $k$ is defined as
\begin{equation}
    u^{(k)}(\mathbf{x}) = \tilde{\mathbf{s}}^{(k)}(\mathbf{x})^\top\boldsymbol{\alpha}^*_k \in \mathbb{R},
    \label{eq:update_function}
\end{equation}
where $\tilde{\mathbf{s}}^{(k)}(\mathbf{x}) = \operatorname{diag}(\boldsymbol{\delta}_k)^{-1}\bigl(\mathbf{s}^{(k)}(\mathbf{x}) - \boldsymbol{\nu}_k\bigr) \in \mathbb{R}^M$. Finally, the model is updated as
\begin{equation}
    F_k(\mathbf{x}) = F_{k-1}(\mathbf{x}) + \eta\, u^{(k)}(\mathbf{x}).
    \label{eq:model_update}
\end{equation}

\subsection{Convergence of the boosting process}

When $\lambda=0$, the candidate vector $\mathbf{u}^{(k)} = \tilde{\mathbf{S}}^{(k)} \boldsymbol{\alpha}^*_k \in \mathbb{R}^n$, with $\big[ \mathbf{u}^{(k)} \big]_i = u^{(k)}(\mathbf{x}_i)$, represents the orthogonal projection of $\mathbf{r}_k$ onto the semantic subspace $\mathcal{H}_k \subseteq \mathbb{R}^n$, such that
\begin{equation}
    \mathcal{H}_k = \{ \mathbf{v} \in \mathbb{R}^n : \mathbf{v} = \tilde{\mathbf{S}}^{(k)} \boldsymbol{\alpha}, \, \boldsymbol{\alpha} \in \mathbb{R}^M \},
\end{equation}
\begin{equation}
    \mathbf{u}^{(k)} = \text{proj}_{\mathcal{H}_k}(\mathbf{r}_k).
    \label{eq:proj}
\end{equation}

The residual vector is $\mathbf{r}_{k} = \mathbf{u}^{(k)} + \mathbf{e}_k$, where the approximation error $\mathbf{e}_k$ is orthogonal to $\mathcal{H}_k$ and thus, $\mathbf{e}_k \perp \mathbf{u}^{(k)}$. The learning trajectory of the algorithm can therefore be interpreted as a sequence of discrete jumps in the semantic space.

In the regression case, given that $\mathbf{u}^{(k)}$ is the component of $\mathbf{r}_{k}$ contained in $\mathcal{H}_k$, the norm of the training residual is reduced, using orthogonality $\mathbf{e}_k \perp \mathbf{u}^{(k)}$, to
\begin{equation}
    \|\mathbf{r}_{k+1}\|_2^2 = \|\mathbf{r}_{k}\|_2^2 - (2\eta - \eta^2) \|\mathbf{u}^{(k)}\|_2^2.
\end{equation}

The convergence condition $\|\mathbf{r}_{k+1}\|_2 < \|\mathbf{r}_{k}\|_2$ implies that $g(\eta) = 2\eta - \eta^2$, with roots at $\eta=0$ and $\eta=2$, must be positive, which proves that the error decreases monotonically in the interval $(0, 2)$, provided that $\|\mathbf{u}^{(k)}\|_2>0$. The orthogonal projection with the smallest residual is obtained at the vertex of $g(\eta)$, which is located at $\eta=1$, but in practice a rate of $\eta \le 1$ is employed to avoid overfitting and improve the generalization capability of the model.

For the case of binary classification, the residual is non-linear in $\eta$ due to $\sigma(\cdot)$, so instead of analyzing the norm of the residual, the decrease in the log-loss is considered as
\begin{equation}
    \mathcal{L}(F) = -\frac{1}{n}\sum_{i=1}^n \Bigl[y_i \log \sigma(F(\mathbf{x}_i)) + (1-y_i)\log\bigl(1-\sigma(F(\mathbf{x}_i))\bigr)\Bigr].
    \label{eq:logloss}
\end{equation}

Defining $h(\eta) = \mathcal{L}(F_{k-1} + \eta\, u^{(k)})$, the decrease condition $\mathcal{L}(F_k) < \mathcal{L}(F_{k-1})$ is equivalent to $h(\eta) < h(0)$. Applying the chain rule to each term of the sum in \eqref{eq:logloss} yields
\begin{equation}
    h'(\eta) = -\frac{1}{n}\sum_{i=1}^n r_i(\eta)\, u^{(k)}(\mathbf{x}_i),
\end{equation}
where $r_i(\eta) = y_i - \sigma\bigl(F_{k-1}(\mathbf{x}_i) + \eta\, u^{(k)}(\mathbf{x}_i)\bigr)$. Evaluating at $\eta = 0$, where $r_i(0) = r_{k,i}$, and using the orthogonal projection property $\mathbf{r}_k^\top \mathbf{u}^{(k)} = \|\mathbf{u}^{(k)}\|_2^2$ (valid for $\lambda = 0$), gives
\begin{equation}
    h'(0) = -\frac{1}{n}\mathbf{r}_k^\top\mathbf{u}^{(k)} = -\frac{\|\mathbf{u}^{(k)}\|_2^2}{n}.
    \label{eq:diff_h}
\end{equation}

Differentiating again and bounding with $\sigma'(z) \leq \tfrac{1}{4}\ \forall\, z \in \mathbb{R}$ gives
\begin{equation}
    h''(\eta) = \frac{1}{n}\sum_{i=1}^n \sigma'\bigl(F_{k-1}(\mathbf{x}_i) + \eta\, u^{(k)}(\mathbf{x}_i)\bigr)\,\bigl(u^{(k)}(\mathbf{x}_i)\bigr)^2 ,
    \label{eq:hess_bound}
\end{equation}
\begin{equation}
    h''(\eta) \leq \frac{\|\mathbf{u}^{(k)}\|_2^2}{4n}.
    \label{eq:d2h_cota_sup}
\end{equation}
Applying Taylor's theorem with the Lagrange remainder yields
\begin{equation}
    \mathcal{L}(F_k) \leq \mathcal{L}(F_{k-1}) 
    - \frac{\eta}{n}\!\left(1 - \frac{\eta}{8}\right)\|\mathbf{u}^{(k)}\|_2^2,
    \label{eq:classif_descent}
\end{equation}
which guarantees a reduction in the loss for $\eta \in (0, 8)$ provided that $\|\mathbf{u}^{(k)}\|_2 > 0$. The maximum descent of the bound is reached at $\eta = 4$, the vertex of $f(\eta) = \eta(1 - \eta/8)$. However, this is a sufficient, but not necessary, condition, because the actual curvature of $h$ is less than that described in \eqref{eq:d2h_cota_sup}, since $\sigma'(z) = \tfrac{1}{4}$ only when $F_{k}(\mathbf{x}_i) = 0\ \forall\, i$.

On the other hand, when $\lambda \neq 0$, the Elastic Net penalty acts as an additional constraint which, although it breaks strict orthogonality, is a well-known mechanism for promoting generalization in the model.

\subsection{Adaptive Geometric Early Stopping}\label{subsec:ages}

The objective is to ensure that the updates contribute significantly to the convergence of the model and to prevent overfitting. To achieve this, an early stopping (ES) mechanism is included, whose main novelty is the use of the geometric angle $\theta_k$ between the predictions and the targets as a convergence metric, combined with a dynamic tolerance that adapts to the learning speed observed during the initial levels. Unlike traditional stopping criteria \citep{prechelt1998early} based on the validation error, whose range depends on the scale of the problem, $\theta_k$ is bounded within the interval [$0, \pi$], which allows defining relative and adaptive thresholds independent of the dataset. This strategy distinguishes between stagnation due to slow improvement and degradation due to overfitting. It allows reverting the model to the best previous state. Angular quantities have previously been used to design angle-driven geometric semantic variation operators in GP \citep{chen2019angle_gs_operators}; here, $\theta_k$ is used instead as a bounded early-stopping criterion on the validation predictions, with adaptive tolerance and rollback.

The geometric alignment is calculated on a validation set $(\mathbf{X}_{val}, \mathbf{y}_{val})$. At iteration $k$, the angle $\theta_k$ is defined by
\begin{equation}
    \theta_k = \arccos \left( \frac{\hat{\mathbf{y}}_{val}^{(k)} \cdot \mathbf{y}_{val}}{\|\hat{\mathbf{y}}_{val}^{(k)}\|_2 \|\mathbf{y}_{val}\|_2} \right).
    \label{eq:theta}
\end{equation}

For regression, $\hat{\mathbf{y}}_{val}^{(k)}$ is the prediction vector at level $k$, while in classification, it is the probability vector obtained by applying the sigmoid activation function to the output. This allows the evaluation of its geometric alignment against the target vector $\mathbf{y}_{val}$ without modifying the angle computation.

Geometrically, the algorithm seeks to minimize $\theta_k$ toward $0$. The stopping mechanism consists of a calibration phase and two interruption criteria. First, the initial $W$ levels are used to establish a baseline for the convergence speed. The average absolute change of the angle in this window is calculated to establish a dynamic tolerance
\begin{equation}
    \tau = \gamma \cdot \left( \frac{1}{W-1} \sum_{k=2}^{W} |\Delta\theta_k| \right),
    \label{eq:tolerance}
\end{equation}
\noindent where $\Delta\theta_k = \theta_k - \theta_{k-1}$ and $\gamma \in (0, 1]$ is a percentage scaling factor. Subsequently, the training is stopped due to convergence if the absolute value of the geometric improvement is less than the calculated tolerance for $P_{c}$ consecutive iterations:
\begin{equation}
    |\Delta\theta_i| < \tau \quad \forall i \in \{k-P_{c}+1, \dots, k\}.
    \label{eq:convergence}
\end{equation}

On the other hand, the training is interrupted due to degradation if the slope $\beta_k$, calculated via OLS over a time window of length $L$, is greater than a threshold $\beta_{tol}$ for $P_{d}$ iterations:
\begin{equation}
    \beta_k = \frac{\sum_{i=1}^L (i - \bar{i})(\theta_{k-L+i} - \bar{\theta})}{\sum_{i=1}^L (i - \bar{i})^2},
    \label{eq:slope}
\end{equation}
\noindent where $L$ denotes the length of the time window, $\theta_{k-L+i}$ is the angle at the $i$-th time step within the window, and $\bar{\theta}$ is the average of the angles recorded in that same window.

At termination, the explicit model is formed by the addition of the $K^*$ retained levels:
\begin{equation}
    F_{K^*}(\mathbf{x}) = \sum_{k=1}^{K^*} \eta \sum_{m=1}^M \alpha^*_{k,m} \frac{T_m^{(k)}(\mathbf{x}) - \nu_{k,m}}{\delta_{k,m}},
    \label{eq:final_inference}
\end{equation}
\noindent where, if the training is interrupted due to convergence, the model retains the state of the current iteration ($K^*=k$). However, if it is interrupted due to degradation, the algorithm restores the model to the level that produced the minimum validation error ($K^*=k_{best}$), discarding the updates that induced overfitting; finally, if the mechanism is not triggered at any point, all levels are retained ($K^*=K$).

\subsection{Taxonomic Classification}

Table \ref{tab:stork_atomic_categories} presents an adaptation of the taxonomy of \citet{stork2020new} to describe the behavior of a search algorithm, where the original categories have been described in greater detail to allow for a more precise characterization of each algorithm. This framework classifies optimization algorithms into four functional components: Initialization, Generation, Selection, and Control. Subsequently, Table \ref{table:stork_classification} assigns a label from each dimension to a selection of state-of-the-art methods.

Based on Table \ref{table:stork_classification}, RSB is characterized by a multi-stage generation (\texttt{Gen\_MultiStage}), since it separates the random creation of trees (\texttt{Init\_Set\_Rand}) from the optimization of their coefficients through Elastic Net. The optimization of the coefficients is a deterministic process, the selection component is therefore classified as \texttt{Sel\_Func\_Det}. Finally, the algorithm corresponds to a basic control (\texttt{Ctrl\_Basic}), since its regularization parameters are fixed.

\begin{table*}[t]
    \centering
    \caption{Definition of matrix variables based on the Stork taxonomy.}
    \label{tab:stork_atomic_categories}
    \scriptsize
    \resizebox{\textwidth}{!}{
        \begin{tabular}{llll}
            \toprule
             \textbf{Component} & \textbf{Label} & \textbf{Stork Category} & \textbf{Technical Description} \\ 
            \midrule
            & \texttt{Init\_Single\_Rand} & Single candidate (random) & Random selection of a single initial candidate within the valid search space. \\ 
            & \texttt{Init\_Single\_Know} & Single candidate (knowledge) & Selection of an initial candidate guided by empirical domain knowledge or a priori evaluation. \\ 
            \textbf{Initialization} & \texttt{Init\_Set\_Rand} & Set of candidates (random) & Initialization of a candidate set through uniform random sampling to favor spatial exploration. \\ 
            & \texttt{Init\_Set\_Know\_DOE} & Set of candidates (knowledge/DOE) & Initialization of a candidate set employing design of experiments (DOE) or prior knowledge of the problem. \\ 
            & \texttt{Init\_Comp\_Rand} & Algorithm components (random) & Initial random selection of search operators or low-level heuristics. \\ 
            & \texttt{Init\_Comp\_Pre} & Algorithm components (predefined) & Initial configuration based on a predefined algorithm architecture and fixed search operators. \\ 
            \midrule
            & \texttt{Gen\_LastBest} & Based on last best & Generation of new candidates using only information from the last best observation. \\ 
            & \texttt{Gen\_Selected} & Based on selected & Generation of new candidates based on a specific subset of previously selected solutions. \\ 
            & \texttt{Gen\_Pop\_Set} & Population (selected set) & New solutions created through population operators such as crossover and mutation. \\ 
            \textbf{Generation} & \texttt{Gen\_Dist\_Model} & Distribution (adaptive model) & Sampling of new candidates from a statistical model (e.g., Gaussian) fitted to the population. \\ 
            & \texttt{Gen\_Surr\_Assist} & Fitted surrogate assistance & Generation where a population or distribution is combined with a locally fitted surrogate model. \\ 
            & \texttt{Gen\_Surr\_Global} & Surrogate (global optimization) & Candidates generated by the direct optimization of a globally fitted surrogate model. \\ 
            & \texttt{Gen\_MultiStage} & Multi-stage & Sequential generation integrating probabilistic and surrogate models, as well as heuristic operators in multiple stages. \\ 
            \midrule
            & \texttt{Sel\_Imp\_Best} & Improving the best & Simple elitist strategy: a new candidate is kept only if it improves the current best. \\ 
            & \texttt{Sel\_Weight\_Const} & Weighted/constrained best & Selection of the best candidate based on weighted objectives or constraint penalties. \\ 
            & \texttt{Sel\_Func\_Det} & Selection function (deterministic) & Use of a fixed mathematical rule to choose candidates for the next iteration. \\ 
            \textbf{Selection} & \texttt{Sel\_Func\_Prob} & Selection function (probabilistic) & Selection using stochastic methods such as tournament, rank, or roulette. \\ 
            & \texttt{Sel\_Surr\_Assist} & Surrogate-assisted selection & Use of a surrogate to filter or preselect candidates before an expensive evaluation. \\ 
            & \texttt{Sel\_Infill\_Crit} & Infill criteria (surrogate) & Selection based on acquisition functions such as Expected Improvement or Confidence Bounds. \\ 
            & \texttt{Sel\_Combined} & Combined set & Selection of a heterogeneous set of candidates based on direct improvement, predictions, and exploratory subsets. \\ 
            \midrule
            & \texttt{Ctrl\_Basic} & Basic control & Fixed initialization and variation parameters with simple or null adaptive dynamics. \\ 
            & \texttt{Ctrl\_Advanced} & Advanced control & Use of adaptive variation step sizes and specific parameters of the selection function. \\ 
            \textbf{Control} & \texttt{Ctrl\_Soph\_Pop} & Sophisticated (Population) & Includes self-adaptation, online control, and diversity management in populations. \\ 
            & \texttt{Ctrl\_Soph\_Surr} & Sophisticated (Surrogate) & Management of sampling methods, surrogate selection, and optimizer selection in the surrogate model. \\ 
            & \texttt{Ctrl\_Most\_Soph} & Most Sophisticated & Hyper-control of each selected algorithm component, typical in automated hybrids. \\ 
            \bottomrule
        \end{tabular}
    }
\end{table*}

\begin{table*}[t]
    \centering
    \caption{Functional classification of components based on the Stork et al. taxonomy.}
    \label{table:stork_classification}
    \scriptsize
    \resizebox{\textwidth}{!}{%
        \begin{tabular}{lp{3.5cm}p{3.5cm}p{3.5cm}p{3.5cm}}
            \toprule
            \textbf{Algorithm} &
            \textbf{Initialization} &
            \textbf{Generation} &
            \textbf{Selection} &
            \textbf{Control} \\ \midrule
            GP & 
            Random population of syntax trees (\texttt{Init\_Set\_Rand}). & 
            Crossover and mutation (\texttt{Gen\_Pop\_Set}). & 
            Tournament selection (\texttt{Sel\_Func\_Prob}). & 
            Fixed population size and operator rates (\texttt{Ctrl\_Basic}). \\ \addlinespace
            MRGP & 
            Random population of syntax trees (\texttt{Init\_Set\_Rand}). & 
            Evolution of trees with internal linear regression (\texttt{Gen\_Pop\_Set}). & 
            Tournament selection (\texttt{Sel\_Func\_Prob}). & 
            Fixed evolutionary parameters (\texttt{Ctrl\_Basic}). 
            \\ \addlinespace
            MEGP & 
            Population mapped to MAP-Elites grid (\texttt{Init\_Set\_Rand}). & 
            Evolutionary operators on grid elites (\texttt{Gen\_Pop\_Set}). & 
            Elitist replacement per grid cell (\texttt{Sel\_Imp\_Best}). & 
            KPCA and cosine similarity mapping (\texttt{Ctrl\_Soph\_Pop}). 
            \\ \addlinespace
            GSGP & 
            Random population of syntax trees (\texttt{Init\_Set\_Rand}). & 
            Geometric semantic crossover and mutation (\texttt{Gen\_Pop\_Set}). & 
            Tournament selection (\texttt{Sel\_Func\_Prob}). & 
            Fixed mutation steps and rates (\texttt{Ctrl\_Basic}). 
            \\ \addlinespace
            SBP-GP &
            Random population of syntax trees (\texttt{Init\_Set\_Rand}). &
            Semantic backpropagation variation (e.g., RDO) (\texttt{Gen\_Pop\_Set}). &
            Tournament selection (\texttt{Sel\_Func\_Prob}). &
            Fixed population size, generations, and operator rates (\texttt{Ctrl\_Basic}).
            \\ \addlinespace
            FFX & 
            Expansion of basis functions (\texttt{Init\_Set\_Know\_DOE}). & 
            Pathwise Regularized Learning (\texttt{Gen\_MultiStage}). & 
            Nondominated filtering for Pareto-optimal set (\texttt{Sel\_Func\_Det}). & 
            Fixed regularization path logic (\texttt{Ctrl\_Basic}). 
            \\ \addlinespace
            EFS &
            Seeding with original input variables (\texttt{Init\_Set\_Know\_DOE}). &
            Evolutionary feature composition and pathwise LASSO (\texttt{Gen\_MultiStage}). &
            Pathwise regularized feature subset selection (\texttt{Sel\_Func\_Det}). &
            Fixed operator set, correlation filter, and stop criteria (\texttt{Ctrl\_Basic}).
            \\ \addlinespace
            KP & 
            Random initial team of experts (\texttt{Init\_Set\_Rand}). & 
            Collaborative evolution of team members (\texttt{Gen\_Pop\_Set}). & 
            $p$-value and coefficient threshold testing (\texttt{Sel\_Func\_Det}). & 
            Restart and Expansion Factor on stagnation (\texttt{Ctrl\_Soph\_Pop}). 
            \\ \addlinespace
            RC & 
            Random non-linear RNN reservoir (\texttt{Init\_Single\_Rand}). & 
            Direct analytical computation, no iterative generation (N/A). & 
            Ridge Regression or Least Squares (\texttt{Sel\_Func\_Det}). & 
            Fixed global hyperparameters such as spectral radius (\texttt{Ctrl\_Basic}). \\ \addlinespace
            SR-Forest & 
            Random initial population (\texttt{Init\_Set\_Rand}). & 
            Crossover and guided mutation (\texttt{Gen\_Pop\_Set}). & 
            Lexicase Selection on randomized cases (\texttt{Sel\_Func\_Prob}). & 
            Mutation and Multi-fidelity evaluation (\texttt{Ctrl\_Soph\_Pop}). \\ \addlinespace
            SyRBo &
            Random population of syntax trees (\texttt{Init\_Set\_Rand}). &
            Evolution of trees to fit gradient residuals (\texttt{Gen\_Pop\_Set}). &
            Tournament selection during evolution (\texttt{Sel\_Func\_Prob}). &
            Fixed evolutionary and boosting parameters (\texttt{Ctrl\_Basic}).
            \\ \addlinespace
            SBR &
            Random sampling of semantic trees (\texttt{Init\_Set\_Rand}). &
            MSGP with semantic backpropagation (\texttt{Gen\_MultiStage}). &
            Elitist acceptance by MSE and size on validation (\texttt{Sel\_Imp\_Best}). &
            Fixed MSGP and scaling parameters (\texttt{Ctrl\_Basic}).
            \\ \addlinespace
            \textbf{RSB} & 
            Random sampling of semantic trees (\texttt{Init\_Set\_Rand}). & 
            Boosting via residual optimization (\texttt{Gen\_MultiStage}). & 
            Deterministic Elastic Net minimization (\texttt{Sel\_Func\_Det}). & 
            Fixed learning rate and algorithm hyperparameters (\texttt{Ctrl\_Basic}). 
            \\ \bottomrule
        \end{tabular}%
    }
\end{table*}

\begin{figure}[!tbp]
    \centering
    \includegraphics[width=\columnwidth]{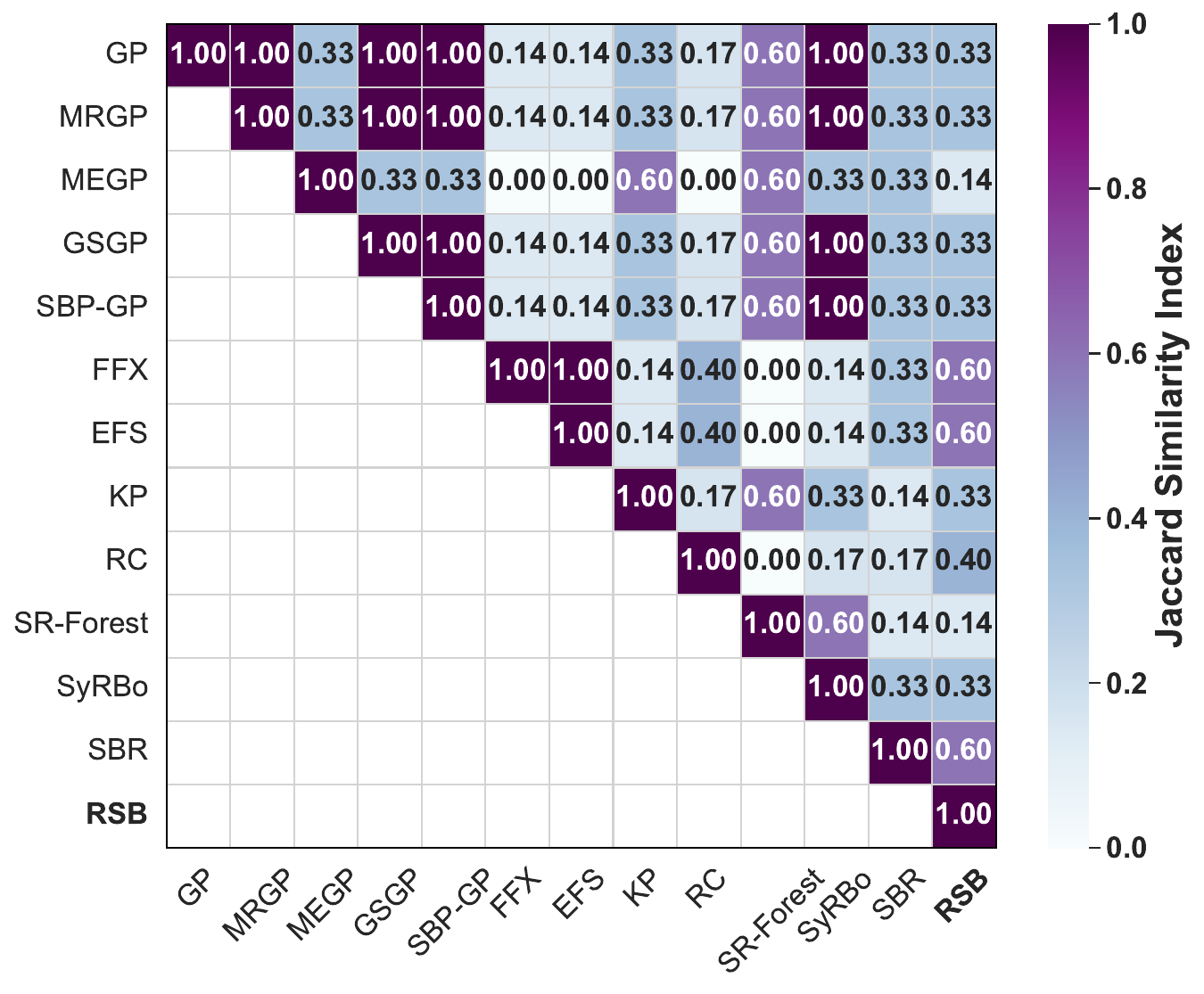}
    \caption{Jaccard similarity matrix quantifying the overlap of algorithmic components across the 13 algorithms in Table~\ref{table:stork_classification}.}
    \label{fig:jaccard_matrix}
\end{figure}

Based on Table \ref{table:stork_classification}, a Jaccard similarity matrix was generated (Figure \ref{fig:jaccard_matrix}) to quantify how related the methods are in terms of their search strategy, considering traditional evolutionary methods (GP, GSGP, MRGP, MEGP, SBP-GP), pathwise-regularized and reservoir-based algorithms (FFX, EFS, RC), and ensemble or boosting approaches (KP, SR-Forest, SyRBo, SBR). In this figure, a similarity of 0.60 is observed between RSB, FFX, EFS and SBR, indicating that, under the Stork taxonomy, they share a similar search strategy. Another method with which RSB shows a notable similarity is RC (0.40); this is due to a common architecture based on stochastic projection and deterministic linear regression.

\subsection{Learning Process Dynamics}

To geometrically illustrate the optimization process of RSB, an illustrative problem was defined with three one-dimensional instances $\mathbf{X}_{tr} = [-1.0, 0.0, 1.0]^\top$ and a target vector $\mathbf{y} = [1.0, 1.5, 3.0]^\top$, using a configuration of $M=2$, $d_{max}=1$, and $\eta=1$. Selecting only two linearly independent trees ensures that the semantic subspace generated at each boosting level is a plane in $\mathbb{R}^3$, which allows for a better visualization of each vector.

\begin{figure*}[!t]
    \centering
    \tikzset{
        lbl/.style={font=\scriptsize\bfseries, inner sep=1pt},
        step_label/.style={lbl, blue, anchor=south}
    }
    \begin{subfigure}[b]{0.4\textwidth}
        \centering
        \begin{tikzpicture}
            \node[anchor=south west, inner sep=0] (img) at (0,0) {\includegraphics[width=\linewidth]{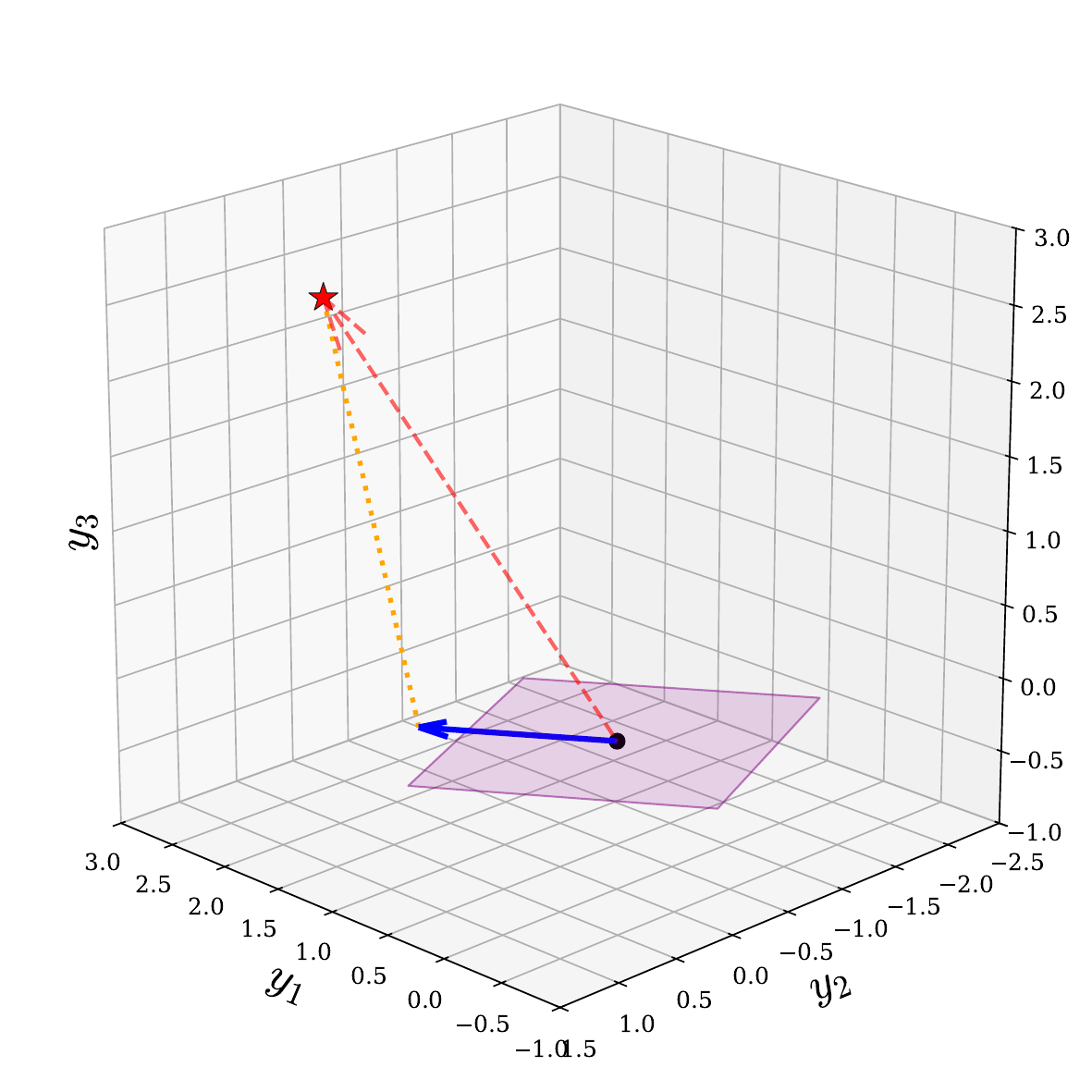}};
            \begin{scope}[x={(img.south east)}, y={(img.north west)}]
                \node[step_label, rotate=-15] at (0.7, 0.85) {1st Boosting Level};
            \end{scope}
        \end{tikzpicture}
    \end{subfigure}
    \hspace{0.3cm}
    \begin{subfigure}[b]{0.4\textwidth}
        \centering
        \begin{tikzpicture}
            \node[anchor=south west, inner sep=0] (img) at (0,0) {\includegraphics[width=\linewidth]{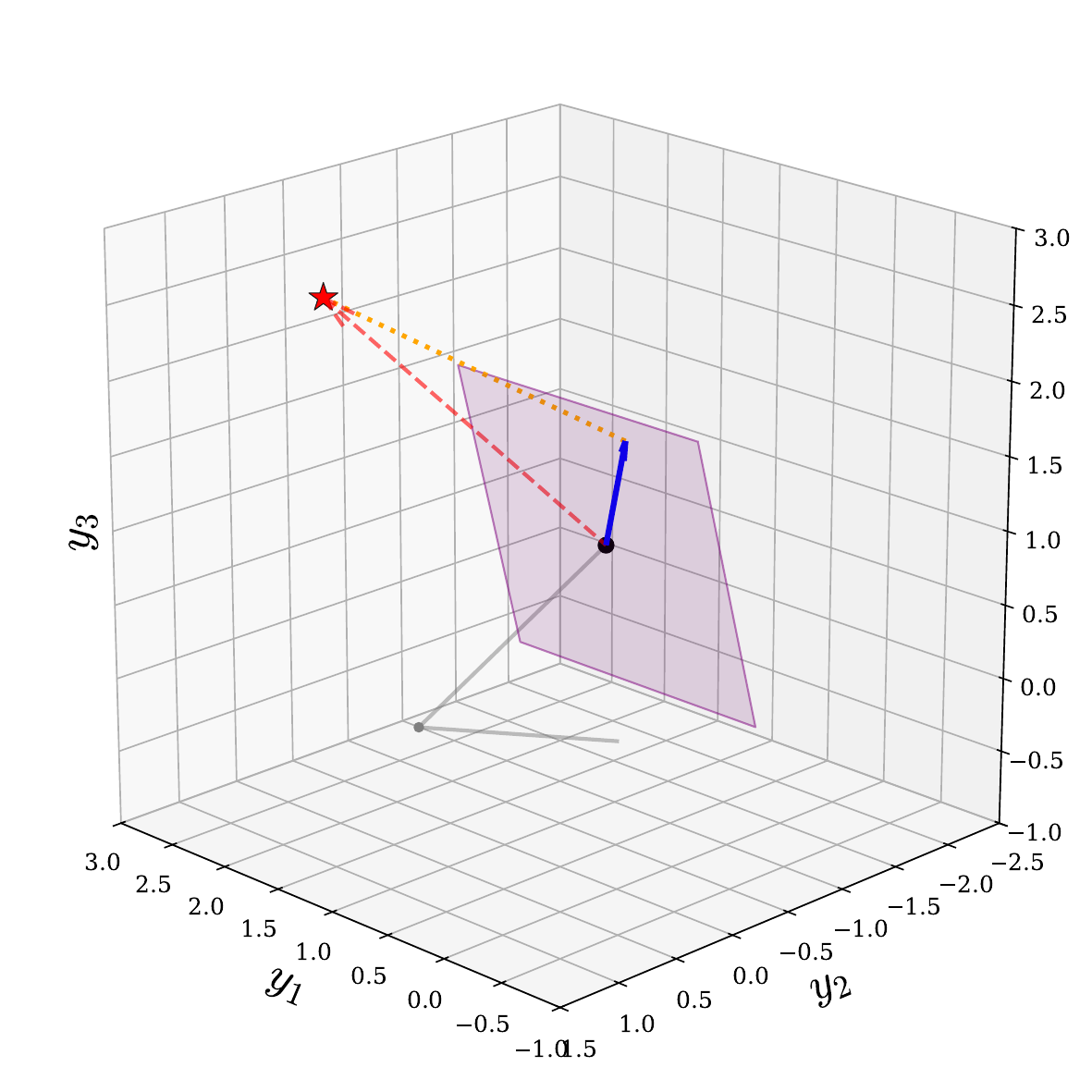}};
            \begin{scope}[x={(img.south east)}, y={(img.north west)}]
                \node[step_label, rotate=-15] at (0.7, 0.85) {3rd Boosting Level};
            \end{scope}
        \end{tikzpicture}
    \end{subfigure}

    \vspace{0.3cm}

    \begin{subfigure}[b]{0.4\textwidth}
        \centering
        \begin{tikzpicture}
            \node[anchor=south west, inner sep=0] (img) at (0,0) {\includegraphics[width=\linewidth]{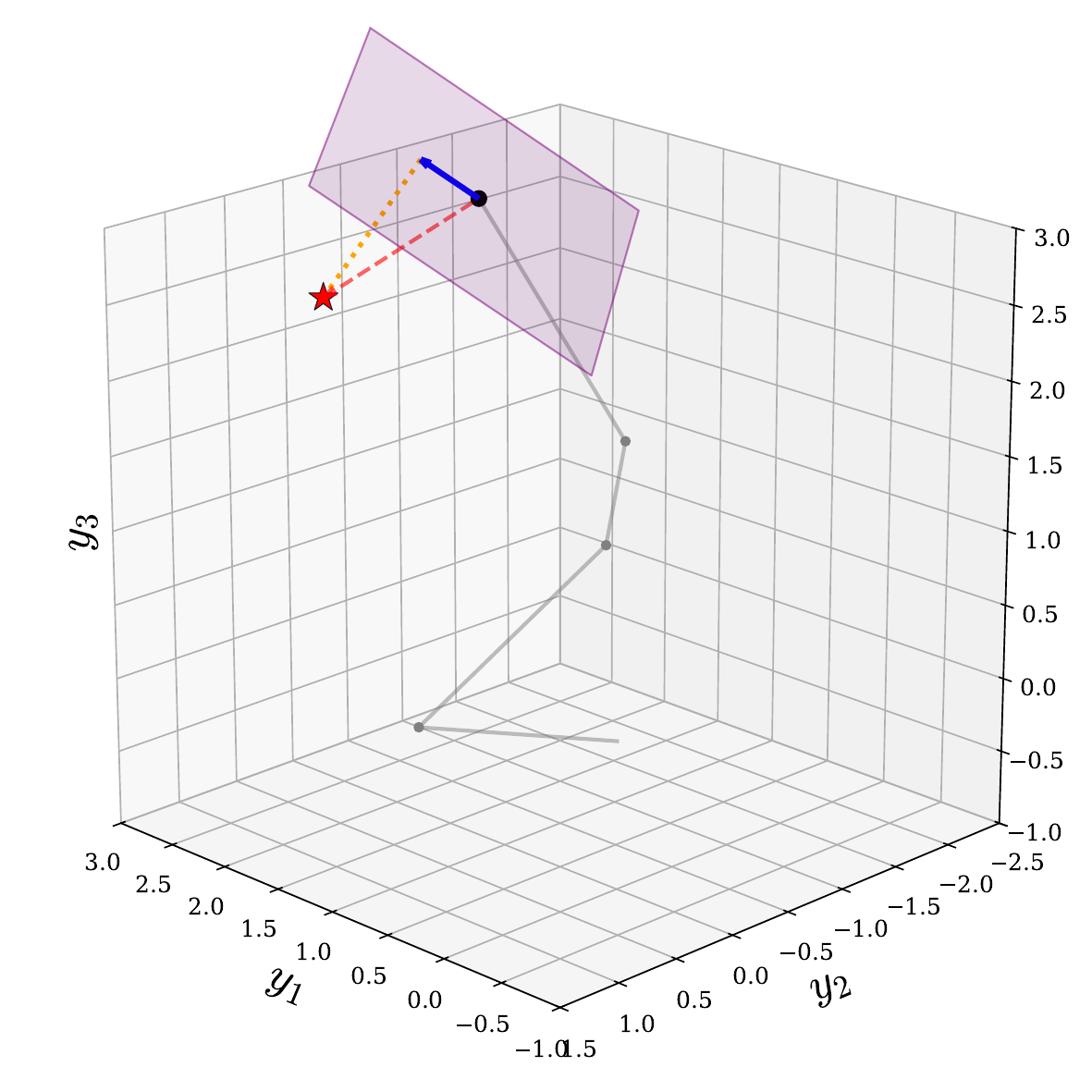}};
            \begin{scope}[x={(img.south east)}, y={(img.north west)}]
                \node[step_label, rotate=-15] at (0.7, 0.85) {5th Boosting Level};
            \end{scope}
        \end{tikzpicture}
    \end{subfigure}
    \hspace{0.3cm}
    \begin{subfigure}[b]{0.4\textwidth}
        \centering
        \begin{tikzpicture}
            \node[anchor=south west, inner sep=0] (img) at (0,0) {\includegraphics[width=\linewidth]{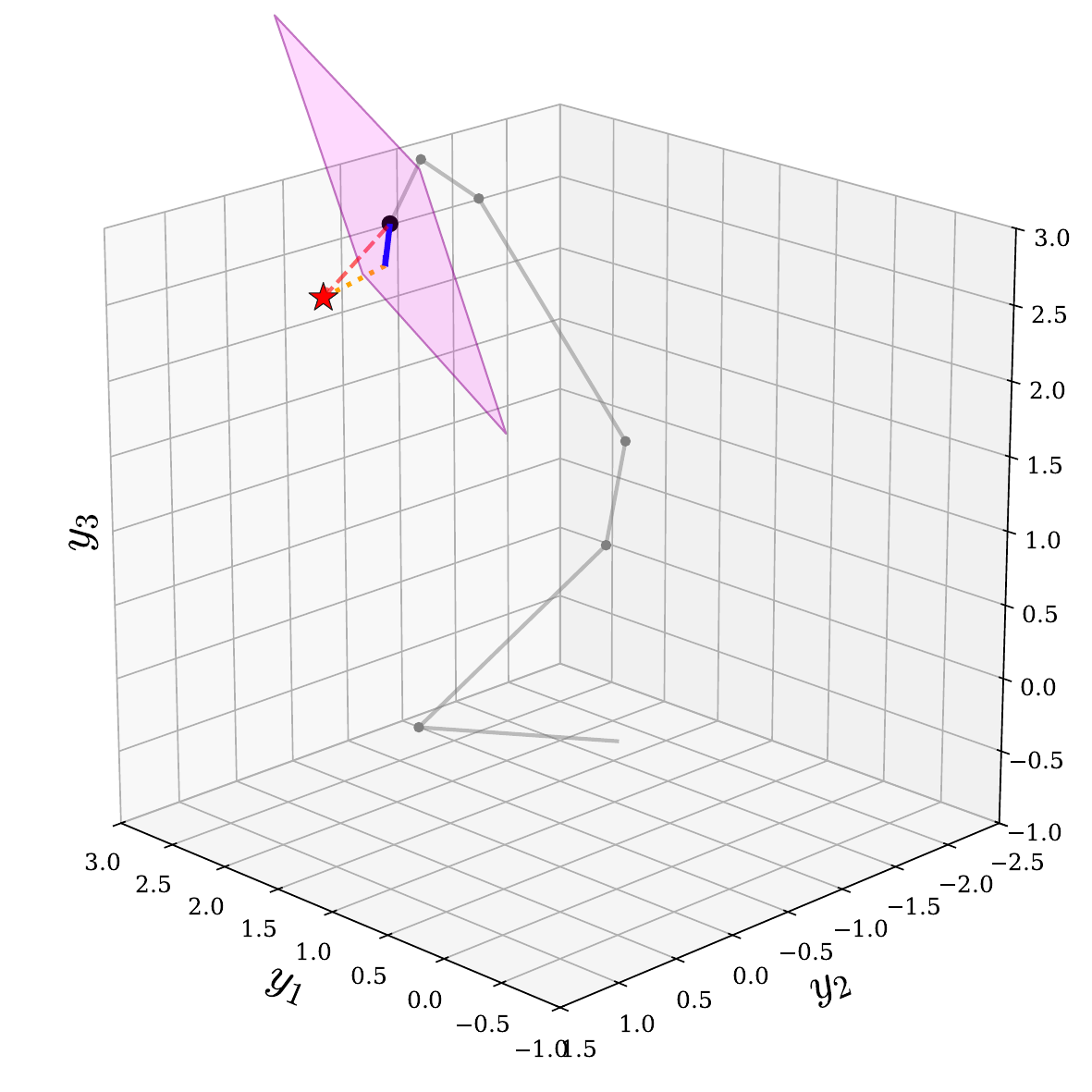}};
            \begin{scope}[x={(img.south east)}, y={(img.north west)}]
                \node[step_label, rotate=-15] at (0.7, 0.85) {7th Boosting Level};
            \end{scope}
        \end{tikzpicture}
    \end{subfigure}

    \vspace{0.3cm}

    \begin{minipage}{1\columnwidth} 
        \centering 
        \scriptsize
        \setlength{\tabcolsep}{5pt}
        \renewcommand{\arraystretch}{1.5} 
        \begin{tabular}{ccccccc} 
            \RefTarget \- $y$ & 
            \RefPosition \- $F_{k-1}$ & 
            \RefSubspace \- $\mathcal{H}_k$ &
            \RefCurrentResidual \- $r_{k}$ &
            \RefPreviousResidual \- $r_{k-1}$ & 
            \RefStep \- $\eta \mathbf{u}^{(k)}$ & 
            \RefHistory \textbf{ History} 
        \end{tabular}
    \end{minipage}

    \caption{Visualization of the geometric dynamics of RSB. The subspace $\mathcal{H}_k$ (magenta surface) is generated at each level. It shows the initialization (black dot), the target $\mathbf{y}$ (red star), the previous residual $\mathbf{r}_{k-1}$ (red dashed line), and the new residual $\mathbf{r}_k$ (orange dotted line) after the projection $\eta \mathbf{u}^{(k)}$ (blue vector). The historical trajectory is traced in gray.}
    \label{fig:geom_dynamics}
\end{figure*}

Figure \ref{fig:geom_dynamics} illustrates the geometric process over seven boosting levels. At each iteration $k$, the semantic matrix $\mathbf{S}^{(k)}$ generates a subspace $\mathcal{H}_k$ that delimits the possible solutions. The model update is performed through the projection $\eta \mathbf{u}^{(k)}$ of the previous residual onto this subspace, which iteratively reduces the magnitude of the new residual $\mathbf{r}_k$, bringing the model prediction closer to the target $\mathbf{y}$.

\section{Experimental Analysis}\label{experimental}

This section analyzes the internal behavior of RSB before comparing its performance with other state-of-the-art methods. The datasets employed are described, followed by a robustness study showing how the algorithm converges globally from different random initializations. Hyperparameter sensitivity is also evaluated using functional analysis of variance (fANOVA), measuring its impact on accuracy and training time. Feature importance is then compared with tree-based ensemble approaches. The section ends with an ablation of regularization and early-stopping criteria, which motivates the configuration used in Section~\ref{benchmarking}.

\subsection{Benchmark Datasets}

To analyze the RSB algorithm in terms of convergence robustness, hyperparameter sensitivity, and feature importance, six real-world regression problems were selected. All problems consist of predicting a single continuous output variable. Table \ref{tab:datasets} summarizes the main characteristics of each dataset, including the number of instances and available input variables.

\begin{table}[t]
    \centering
    \caption{Regression benchmark datasets \citep{concrete,eech,housing,tower,yacht}.}
    \label{tab:datasets}
    \resizebox{\columnwidth}{!}{
        \begin{tabular}{lccl}
            \toprule
            \textbf{Dataset} & \textbf{Features} & \textbf{Samples} & \textbf{Description} \\
            \midrule
            Concrete & 8 & 1030 & High-performance concrete strength \\
            EECooling & 8 & 768 & Building energy cooling load \\
            EEHeating & 8 & 768 & Building energy heating load \\
            Housing & 13 & 506 & Housing prices and air quality \\
            Tower & 25 & 4999 & Unknown real-world relationship \\
            Yacht & 6 & 308 & Yacht hull design and performance \\
            \bottomrule
        \end{tabular}
    }
\end{table}

\subsection{Convergence Robustness}

\begin{figure*}[!tp]
    \centering
    \tikzset{
        lbl/.style={font=\scriptsize\bfseries, inner sep=1pt},
        step_label/.style={lbl, blue, anchor=south}
    }
    \begin{subfigure}[b]{0.32\textwidth}
        \centering
        \begin{tikzpicture}
            \node[anchor=south west, inner sep=0] (img) at (0,0) {\includegraphics[width=\linewidth]{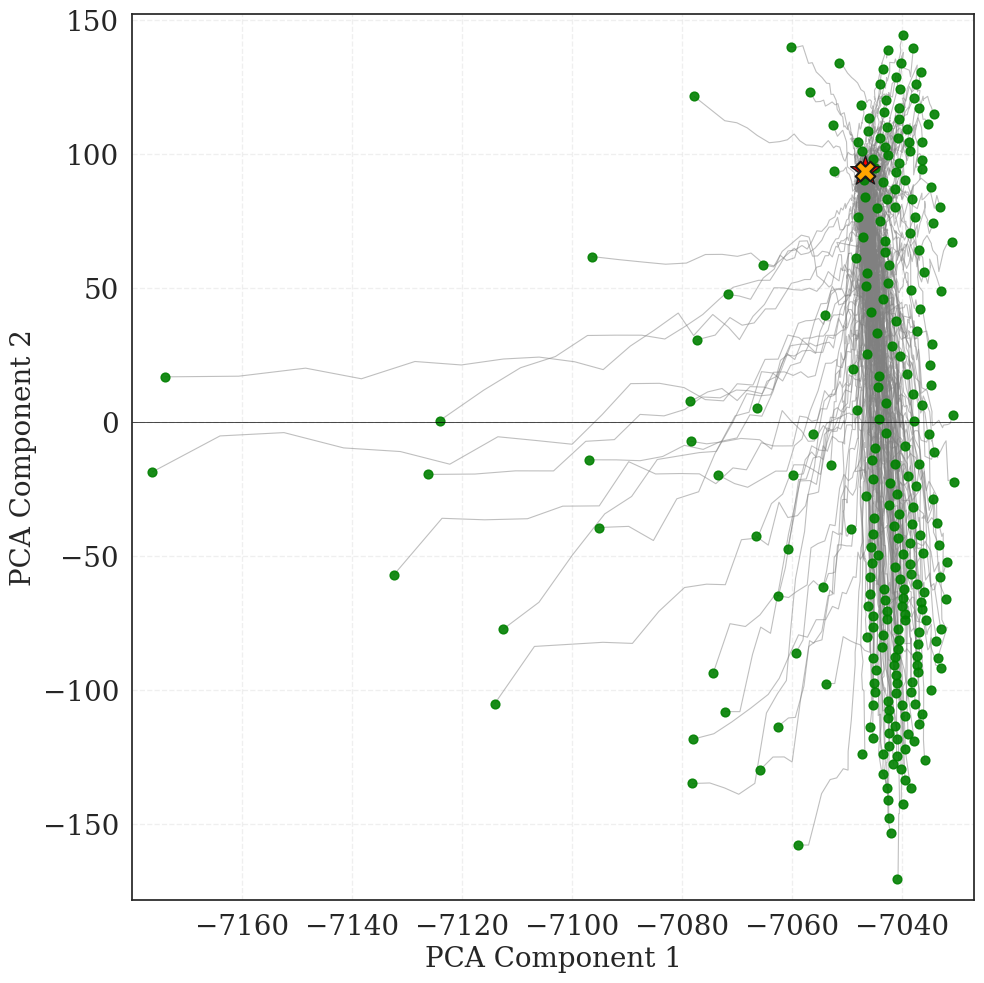}};
            \begin{scope}[x={(img.south east)}, y={(img.north west)}]
                \node[step=0.5, rotate=0] at (0.55, 1.03) {\color{black} \small Concrete Dataset};
            \end{scope}
        \end{tikzpicture}
    \end{subfigure}
    \hfill
    \begin{subfigure}[b]{0.32\textwidth}
        \centering
        \begin{tikzpicture}
            \node[anchor=south west, inner sep=0] (img) at (0,0) {\includegraphics[width=\linewidth]{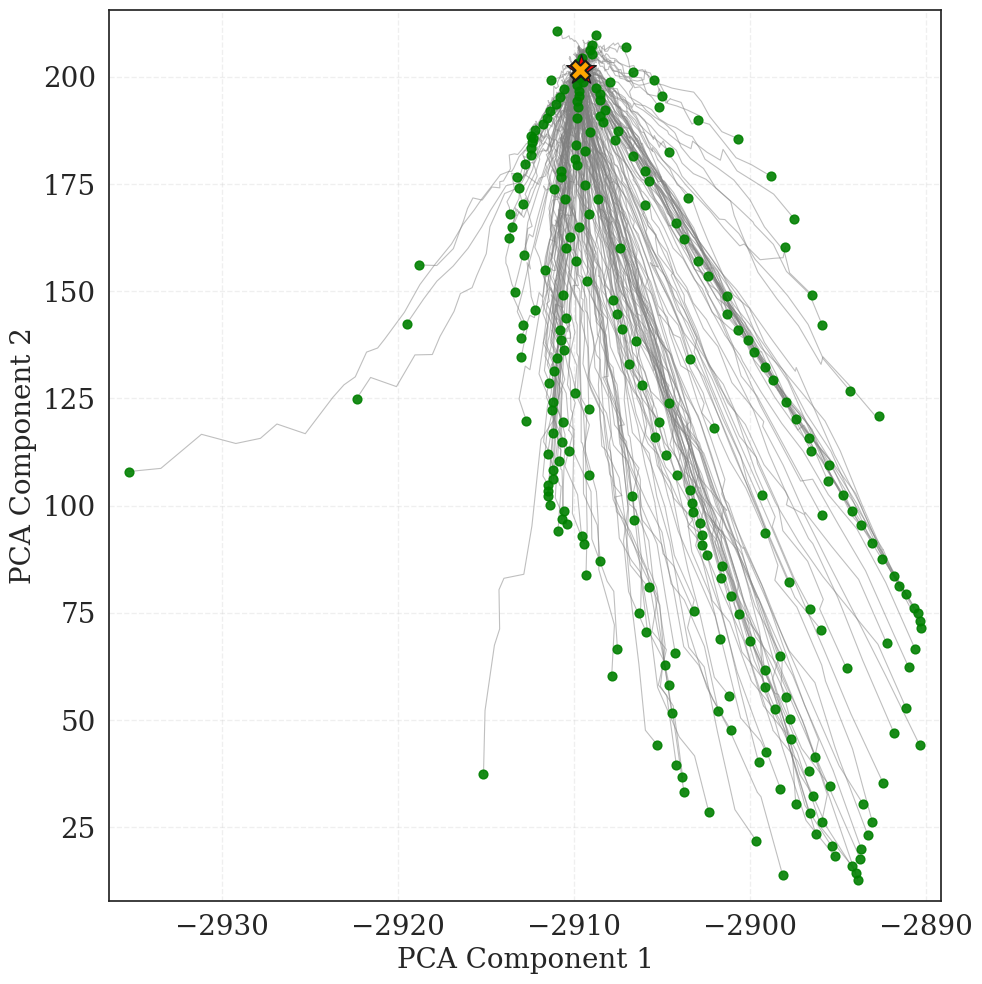}};
            \begin{scope}[x={(img.south east)}, y={(img.north west)}]
                \node[step=0.5, rotate=0] at (0.55, 1.03) {\color{black} \small EECooling Dataset};
            \end{scope}
        \end{tikzpicture}
    \end{subfigure}
    \hfill
    \begin{subfigure}[b]{0.32\textwidth}
        \centering
        \begin{tikzpicture}
            \node[anchor=south west, inner sep=0] (img) at (0,0) {\includegraphics[width=\linewidth]{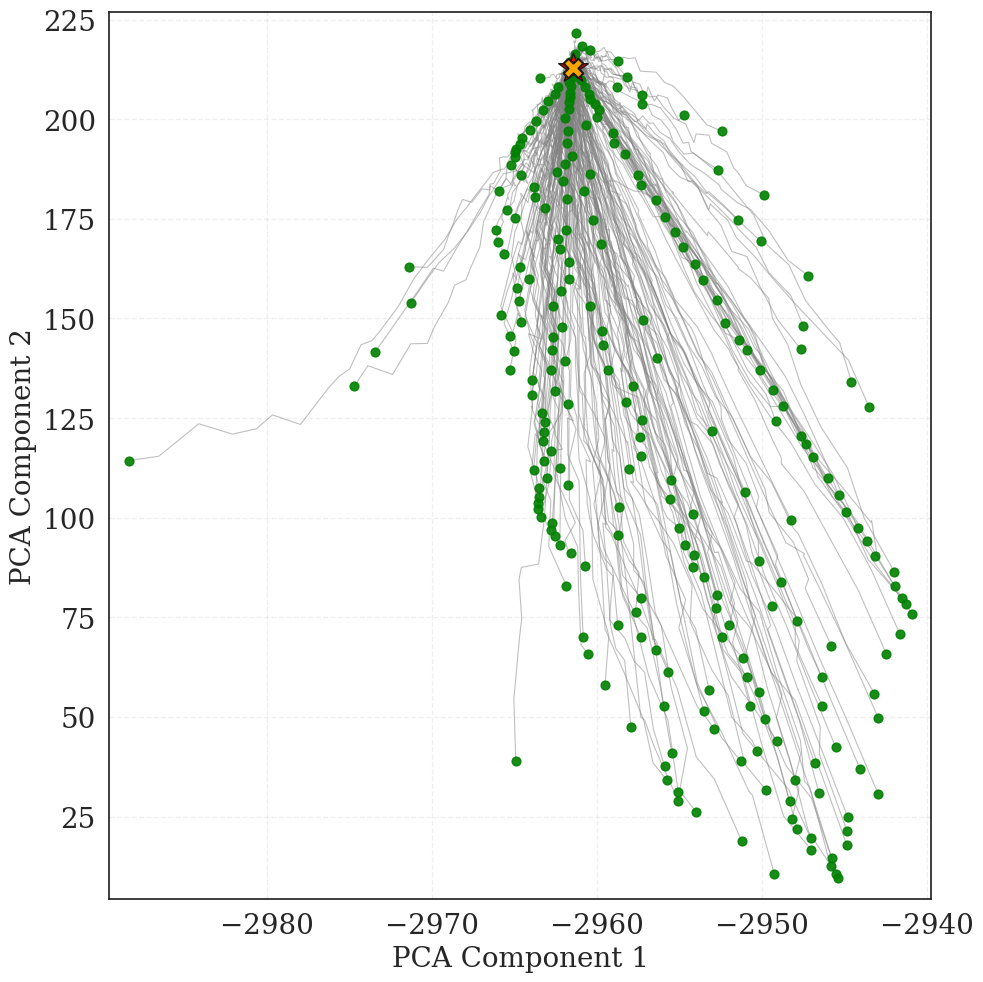}};
            \begin{scope}[x={(img.south east)}, y={(img.north west)}]
                \node[step=0.5, rotate=0] at (0.55, 1.03) {\color{black} \small EEHeating Dataset};
            \end{scope}
        \end{tikzpicture}
    \end{subfigure}

    \vspace{0.3cm}

    \begin{subfigure}[b]{0.32\textwidth}
        \centering
        \begin{tikzpicture}
            \node[anchor=south west, inner sep=0] (img) at (0,0) {\includegraphics[width=\linewidth]{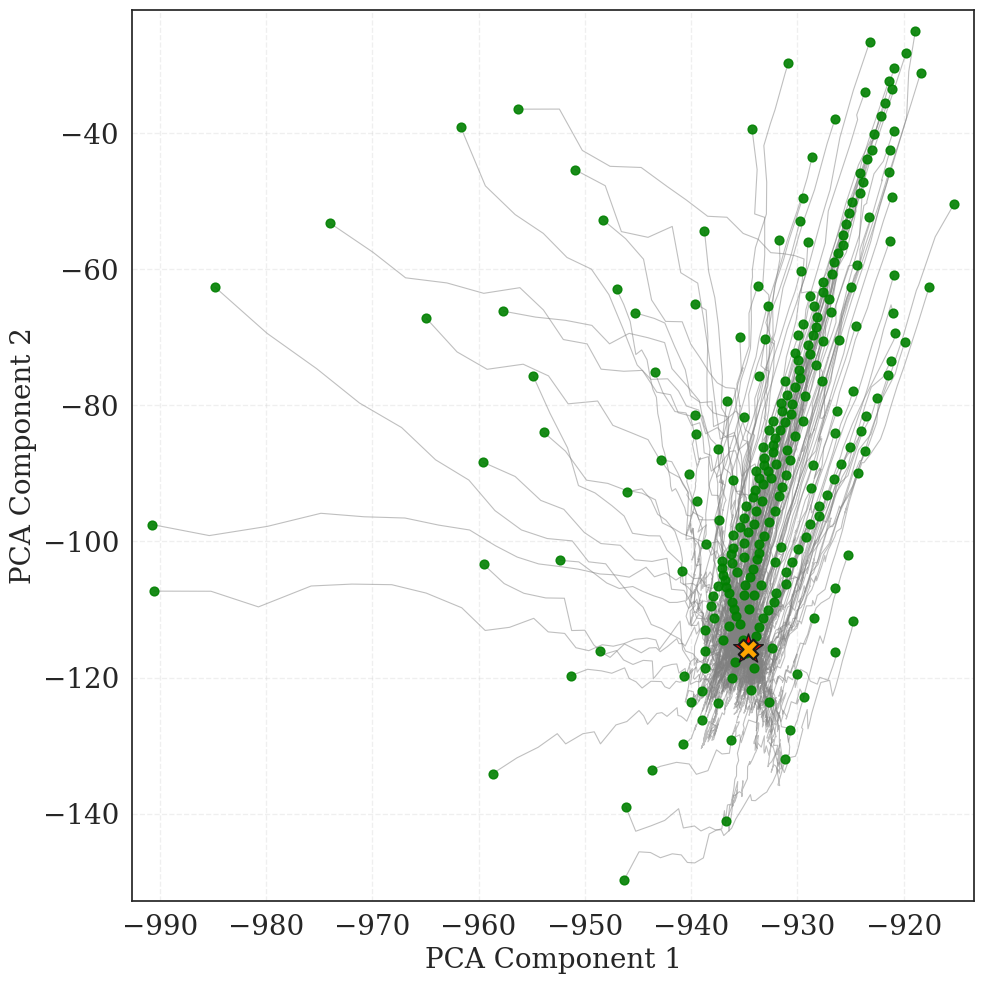}};
            \begin{scope}[x={(img.south east)}, y={(img.north west)}]
                \node[step=0.5, rotate=0] at (0.55, 1.03) {\color{black} \small Housing Dataset};
            \end{scope}
        \end{tikzpicture}
    \end{subfigure}
    \hfill
    \begin{subfigure}[b]{0.32\textwidth}
        \centering
        \begin{tikzpicture}
            \node[anchor=south west, inner sep=0] (img) at (0,0) {\includegraphics[width=\linewidth]{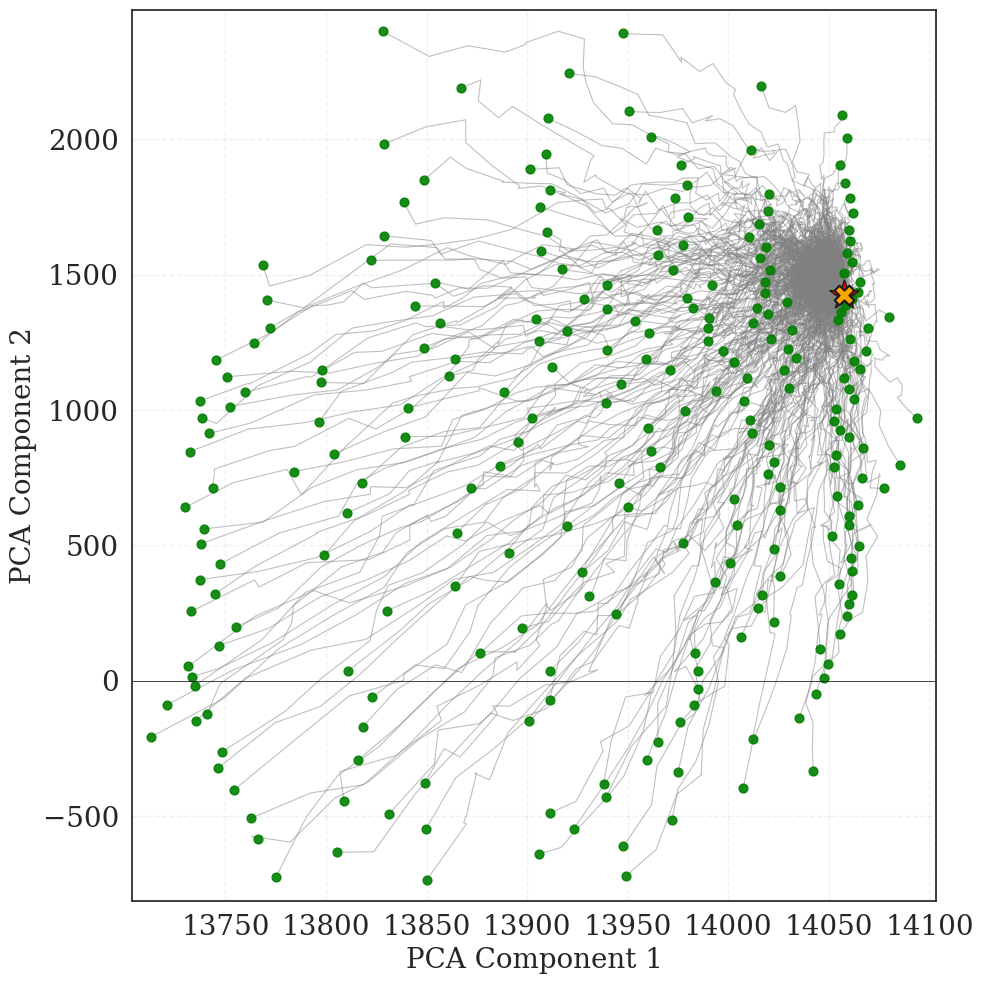}};
            \begin{scope}[x={(img.south east)}, y={(img.north west)}]
                \node[step=0.5, rotate=0] at (0.55, 1.03) {\color{black} \small Tower Dataset};
            \end{scope}
        \end{tikzpicture}
    \end{subfigure}
    \hfill
    \begin{subfigure}[b]{0.32\textwidth}
        \centering
        \begin{tikzpicture}
            \node[anchor=south west, inner sep=0] (img) at (0,0) {\includegraphics[width=\linewidth]{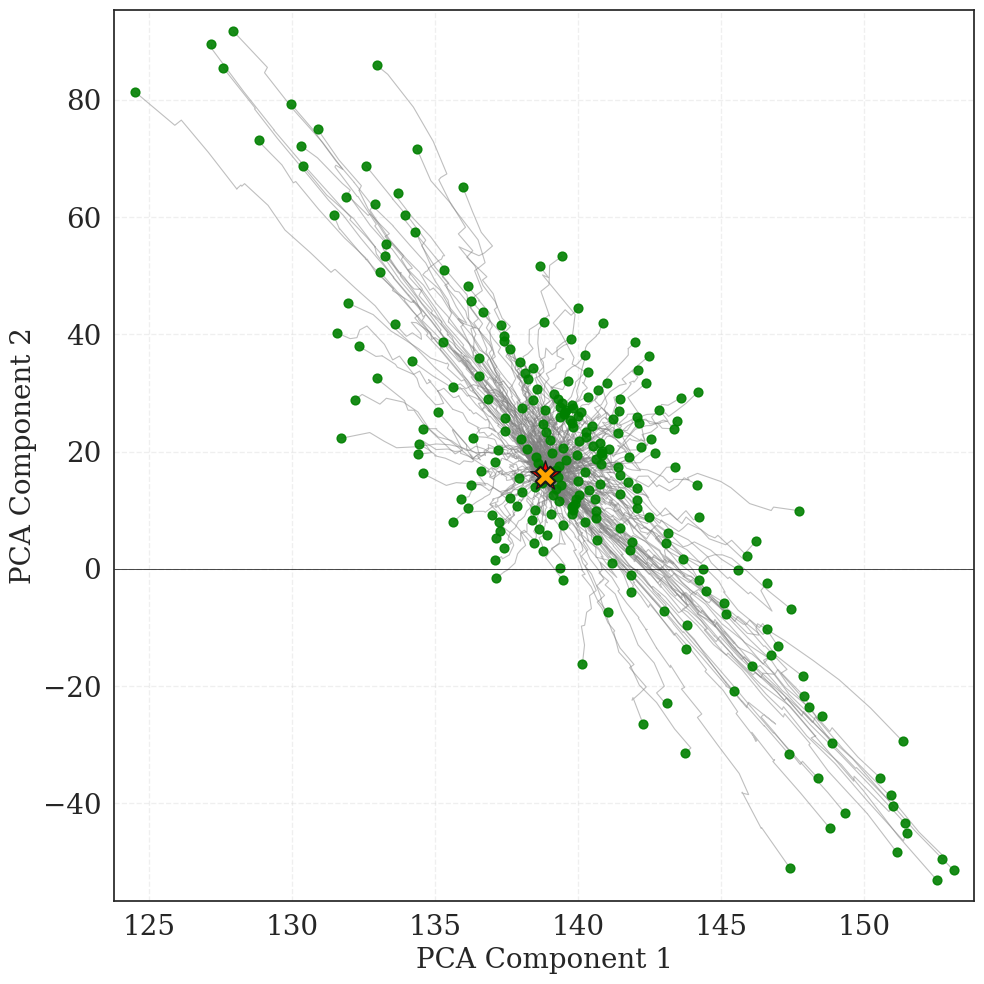}};
            \begin{scope}[x={(img.south east)}, y={(img.north west)}]
                \node[step=0.5, rotate=0] at (0.55, 1.03) {\color{black} \small Yacht Dataset};
            \end{scope}
        \end{tikzpicture}
    \end{subfigure}

    \vspace{0.3cm}

    \begin{minipage}{1\textwidth} 
        \centering 
        \scriptsize
        \renewcommand{\arraystretch}{1.5} 
        \begin{tabular}{cccc} 
            \RefTarget \textbf{ Target} & 
            \RefStart \textbf{ Initialization} & 
            \RefTrajectory \textbf{ Trajectory} & 
            \RefMean \textbf{ Average Convergence Point}
        \end{tabular}
    \end{minipage}

    \caption{Robustness and global convergence analysis projected onto a latent space (PCA-2D) for the datasets described in Table \ref{tab:datasets}. The green dots indicate the initial state of 250 independent simulations, the gray lines their trajectories, and the orange cross ($\times$) denotes the average convergence point with respect to the target (red star).}
    \label{fig:convergence_multiplot}
\end{figure*}

Figure \ref{fig:convergence_multiplot} presents the training dynamics of RSB on the six datasets from Table \ref{tab:datasets}. To facilitate the visual interpretation of vectors in $\mathbb{R}^n$, dimensionality reduction was applied using Principal Component Analysis (PCA), projecting the predictions onto $\mathbb{R}^2$. The green dots represent the initial state of 250 independent simulations, where each state corresponds to the first update $F_1(\mathbf{x}) = \eta \tilde{\mathbf{S}}^{(1)} \boldsymbol{\alpha}_1$ reduced to $\mathbb{R}^2$, with $\eta=0.1$, $M=5$, $d_{max}=3$, and $K=10,000$. Since each initial state consists of random trees and coefficients, there is a wide dispersion of initial conditions. It is observed that, regardless of the initial state, all simulations converge toward a position indistinguishable from the target.

\subsection{Population Size: The $\sqrt{n}$ Scaling Rule}

The selection of hyperparameters depends on the specific problem, but $M$ exhibits the greatest variation. Since $n$ linearly independent vectors are sufficient to generate all of $\mathbb{R}^n$, the number of semantic columns $M$ must be less than $n$ when $\lambda=0$. This restriction is necessary to prevent the subspace from being able to memorize noise, thus preventing overfitting. To achieve this in a scalable manner, $M = m \cdot \sqrt{n}$ is defined; this ensures a sublinear growth where only the constant factor $m$ is adjusted, regardless of the dataset size; when $\lambda>0$, the Elastic Net regularization complements this criterion by directly penalizing the coefficients.

To evaluate how sensitive the RSB algorithm is to its hyperparameter configuration, a surrogate model-based importance analysis was implemented following the methodology of \citet{hutter2014efficient}. This was performed independently for each dataset in Table~\ref{tab:datasets}. An exploration of the search space was carried out using a grid search over the configuration space detailed in Table \ref{tab:hyperparams_importance_values}, where $\text{AQ}(a, b) = a / \sqrt{1 + b^2}$ is the Analytic Quotient operator \citep{ni2013analytic}, without regularization and with early stopping disabled. This resulted in 1500 combinations. The coefficient of determination ($R^2$) and training time were obtained for each using 5-fold cross-validation and 6 runs with different seeds per fold, thus ensuring the statistical reliability of the estimates.

\begin{table}[t]
    \centering
    \caption{Hyperparameter search space and fixed configurations used for the fANOVA analysis.}
    \label{tab:hyperparams_importance_values}
    \resizebox{\columnwidth}{!}{%
    \begin{tabular}{ll}
        \toprule 
        \textbf{Parameter} & \textbf{Values} \\
        \midrule
        $m$ & $\{0.1, 0.25, 0.5, 1, 1.5, 2, 2.5, 3, 3.5, 4, 8, 10\}$ \\
        $K$ & $\{1, 5, 10, 25, 50\}$ \\
        $d_{max}$ & $\{1, 2, 3, 4, 5\}$ \\
        $\eta$ & $\{0.05, 0.1, 0.25, 0.5, 1\}$ \\
        \midrule
        Basic Function Set & $+$, $-$, $\times$, AQ \\
        Regularization & None ($\lambda=0$) \\
        \bottomrule
    \end{tabular}
    }
\end{table}

Subsequently, a Random Forest Regressor model with 100 estimators was trained for each dataset, where the input corresponds to the hyperparameter configuration and the output to the estimated $R^2$ value. From this process, two different analyses were performed: the marginal effect of the factor $m$ and the hyperparameter importance.

\begin{figure*}[!tbp]
    \centering
    \includegraphics[width=0.7\textwidth]{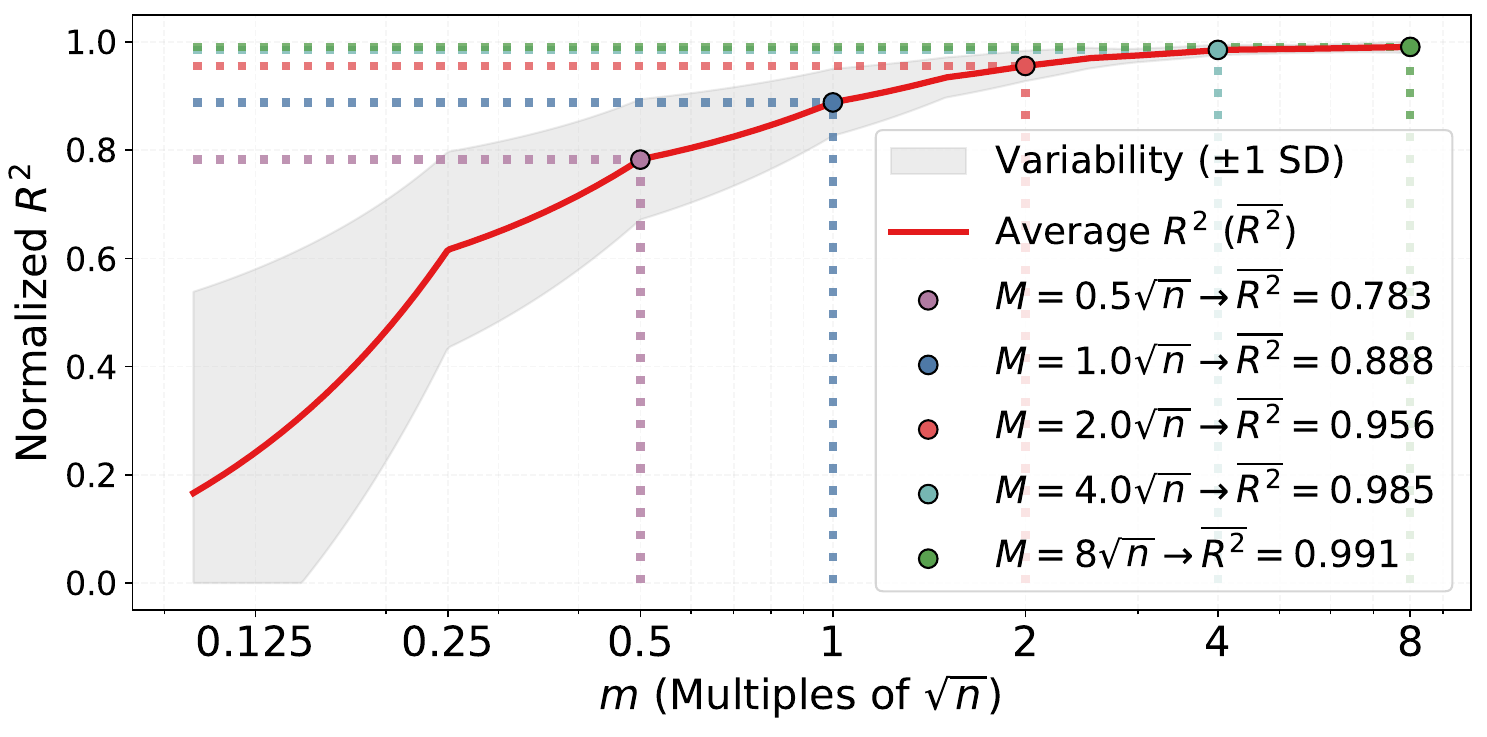}
    \caption{Marginal effect of $m$ on the normalized $R^2$. The solid red line represents the mean response across multiple datasets, while the shaded region denotes the variability ($\pm 1$ SD).}
    \label{fig:m_factor_marginal_effect}
\end{figure*}

Figure \ref{fig:m_factor_marginal_effect} shows the marginal effect of the factor $m$ (obtained using the \texttt{partial\_dependence} function from the \texttt{sklearn.inspection} library in Python). An increasing performance trend is observed as $m$ increases, which confirms that $M \propto \sqrt{n}$ is an effective scaling rule.

\subsection{Hyperparameter Sensitivity}

\begin{figure}[!tbp]
    \centering
    \includegraphics[width=\columnwidth]{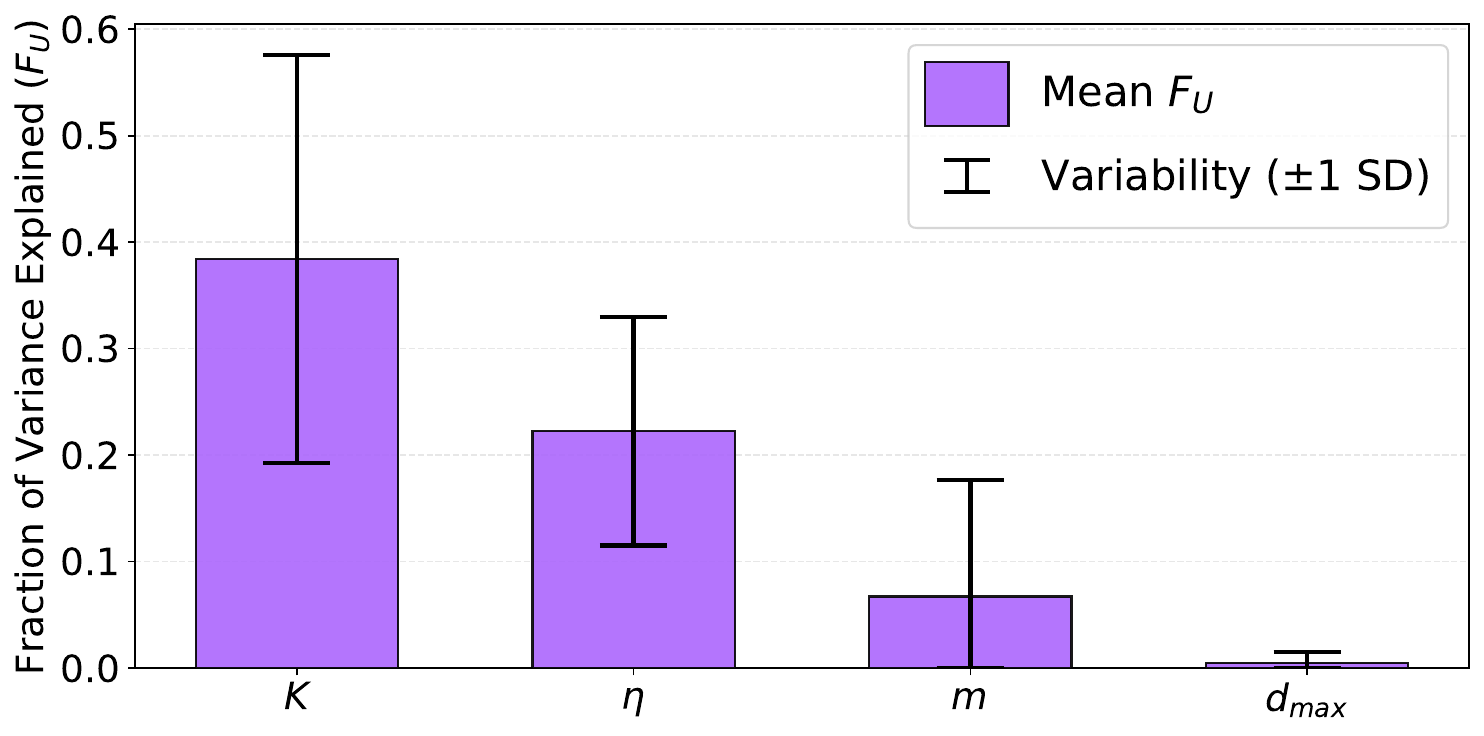}
    \caption{Average hyperparameter importance across datasets using fANOVA (evaluated as the fraction of explained variance, $F_U$). Error bars represent the standard deviation ($\pm 1$ SD).}
    \label{fig:hyp_importance_fanova}
\end{figure}

To understand which hyperparameters have the greatest contribution to the model's performance, fANOVA was used. For each hyperparameter, the fraction of explained variance $F_U=V_U/V$ was calculated, where $V_U$ is the variance of the marginal effect obtained through \texttt{partial\_dependence} and $V$ is the total variance of the surrogate model's predictions. Figure \ref{fig:hyp_importance_fanova} shows the average fraction of variance in $R^2$ explained by each hyperparameter across all the datasets detailed in Table \ref{tab:datasets}. As observed, $K$ and $\eta$ are the most influential factors for the accuracy of the RSB algorithm.

\subsection{Computational Cost}\label{comp_cost}

\begin{figure}[!tbp]
    \centering
    \includegraphics[width=\columnwidth]{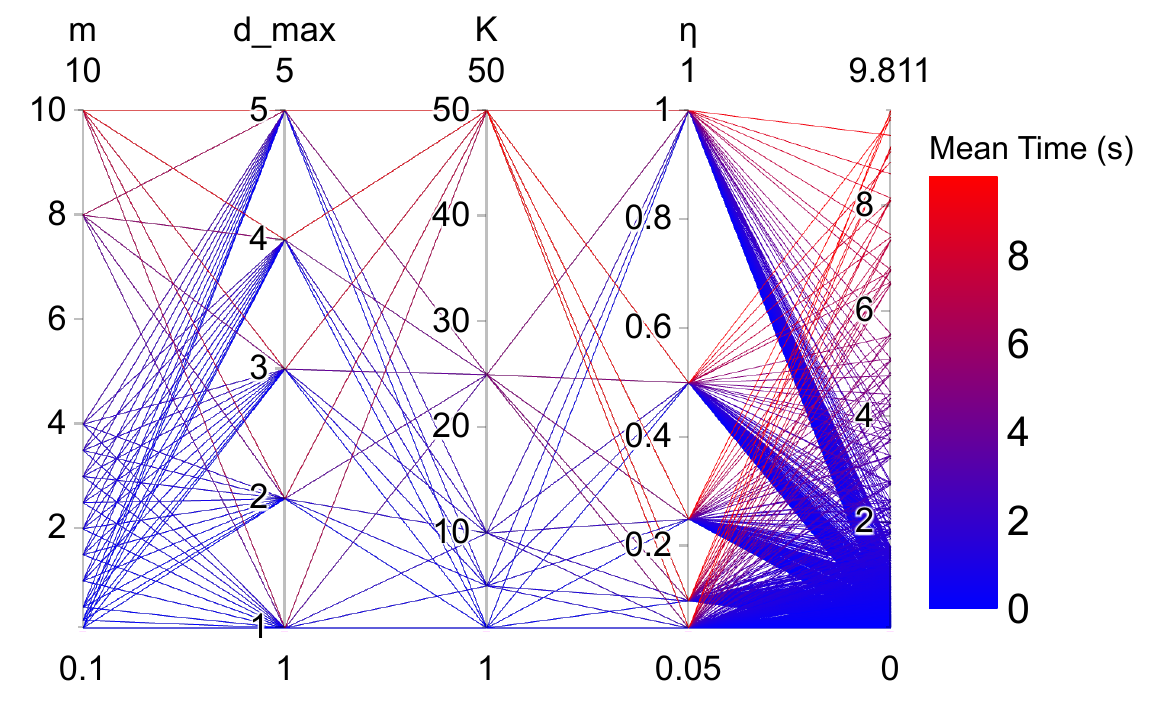}
    \caption{Sensitivity analysis of computational cost for RSB. Each line represents a unique configuration averaged over 30 evaluations (6 runs $\times$ 5 folds). The color scale indicates the execution time in seconds.}
    \label{fig:RSB_time_sensitivity}
\end{figure}

For each of the configurations in Table \ref{tab:hyperparams_importance_values}, the training time was calculated. The experiments were conducted on a machine with Windows 10 Pro, AMD Ryzen 7 5800X at 3.8 GHz (8 cores/16 threads), 64 GB of RAM, and a 1 TB NVMe SSD, without the use of a GPU. Using these times, Figure \ref{fig:RSB_time_sensitivity} was generated; this shows a parallel coordinates plot, used to visualize the relationship between the configurations and the execution time. It can be observed that specific combinations (mainly high values of $m$ and $K$) significantly increase the computational time.

Table \ref{tab:correlation_time_analysis} shows the relationship between each hyperparameter and the training time, using Pearson's and Spearman's correlation coefficients. The analysis shows that $m$ and $K$ have the highest correlation values, followed by $d_{max}$. On the other hand, $\eta$ has a very low value, indicating that it does not contribute much to the computational cost.

\begin{table}[t]
    \centering
    \caption{Statistical correlation analysis of hyperparameters with respect to execution time.}
    \label{tab:correlation_time_analysis}
    \begin{tabular}{lp{2cm}p{2.1cm}}
          \toprule
          \textbf{Parameter} & \textbf{Pearson ($r$)} & \textbf{Spearman ($\rho$)} \\
          \midrule
          $m$ & $0.315998$ & $0.607864$ \\
          $K$ & $0.251069$ & $0.618799$ \\
          $d_{max}$ & $0.070535$ & $0.203032$ \\
          $\eta$ & $-0.000057$ & $-0.000133$ \\
          \bottomrule
    \end{tabular}
\end{table}

\subsection{Feature Importance}

To analyze how much each input variable influences the model's output, a feature importance analysis was implemented alongside RSB training. At each boosting level $k$, the decrease in training Mean Squared Error (MSE) on the current residual after the level update is computed. That decrease is distributed among the $M$ trees at that level in proportion to $|\alpha^*_{k,i}|$ (normalized over the trees).

However, since each tree can contain more than one variable, each variable is randomly permuted on the training set while the others are held fixed, the tree is reevaluated, and the mean squared difference between the baseline and permuted tree outputs is measured. These within-tree values are normalized to sum to one. This calculation was performed across 30 independent runs to obtain statistically robust results.

\begin{figure*}[!tp]
    \centering
    \tikzset{
        lbl/.style={font=\scriptsize\bfseries, inner sep=1pt},
        step_label/.style={lbl, blue, anchor=south}
    }
    \begin{subfigure}[b]{0.47\textwidth}
        \centering
        \begin{tikzpicture}
            \node[anchor=south west, inner sep=0] (img) at (0,0) {\includegraphics[width=\linewidth]{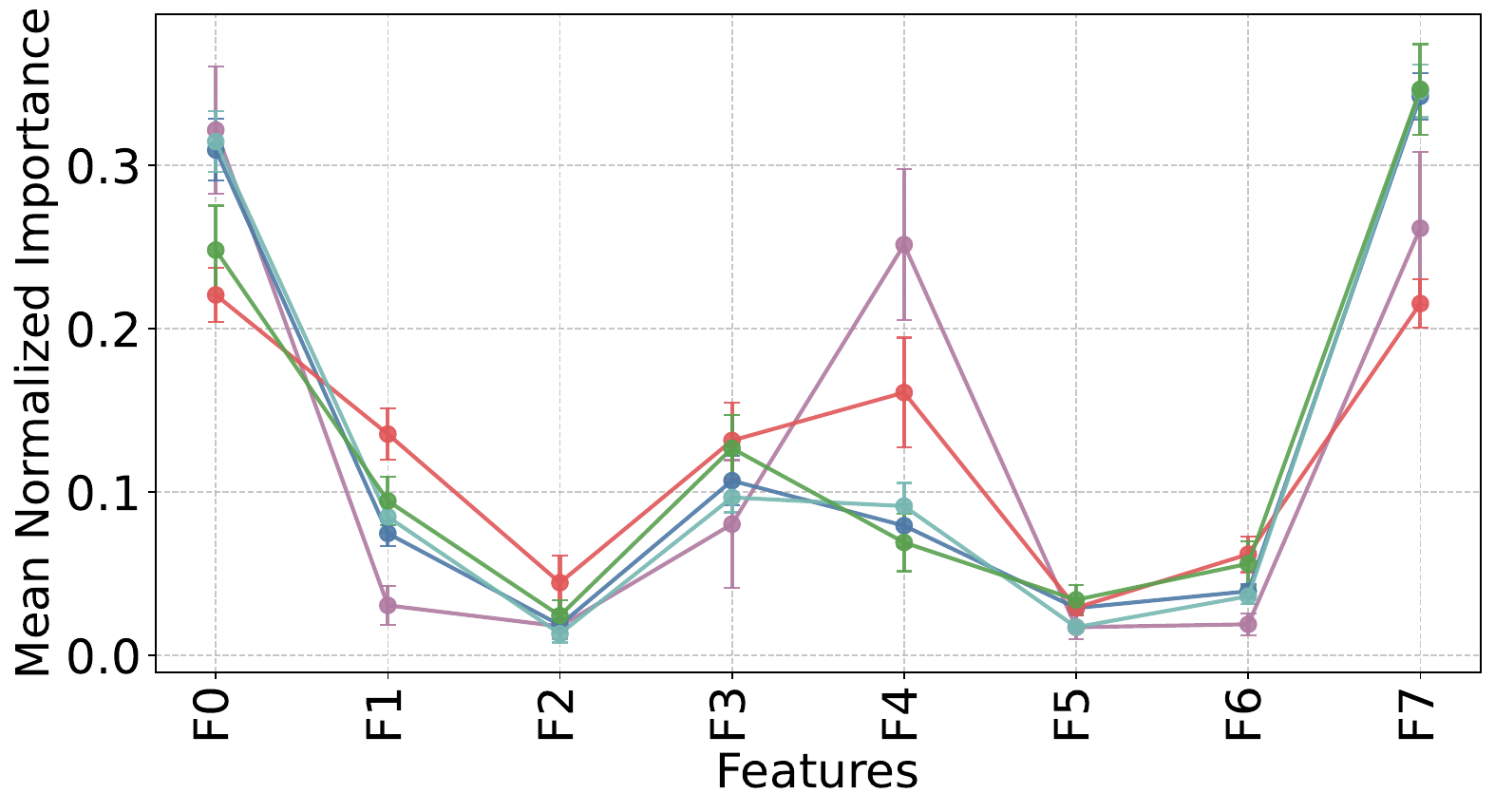}};
            \begin{scope}[x={(img.south east)}, y={(img.north west)}]
                \node[step=0.5, rotate=0] at (0.55, 0.9) {\color{black} \small Concrete Dataset};
            \end{scope}
        \end{tikzpicture}
    \end{subfigure}
    \hfill
    \begin{subfigure}[b]{0.47\textwidth}
        \centering
        \begin{tikzpicture}
            \node[anchor=south west, inner sep=0] (img) at (0,0) {\includegraphics[width=\linewidth]{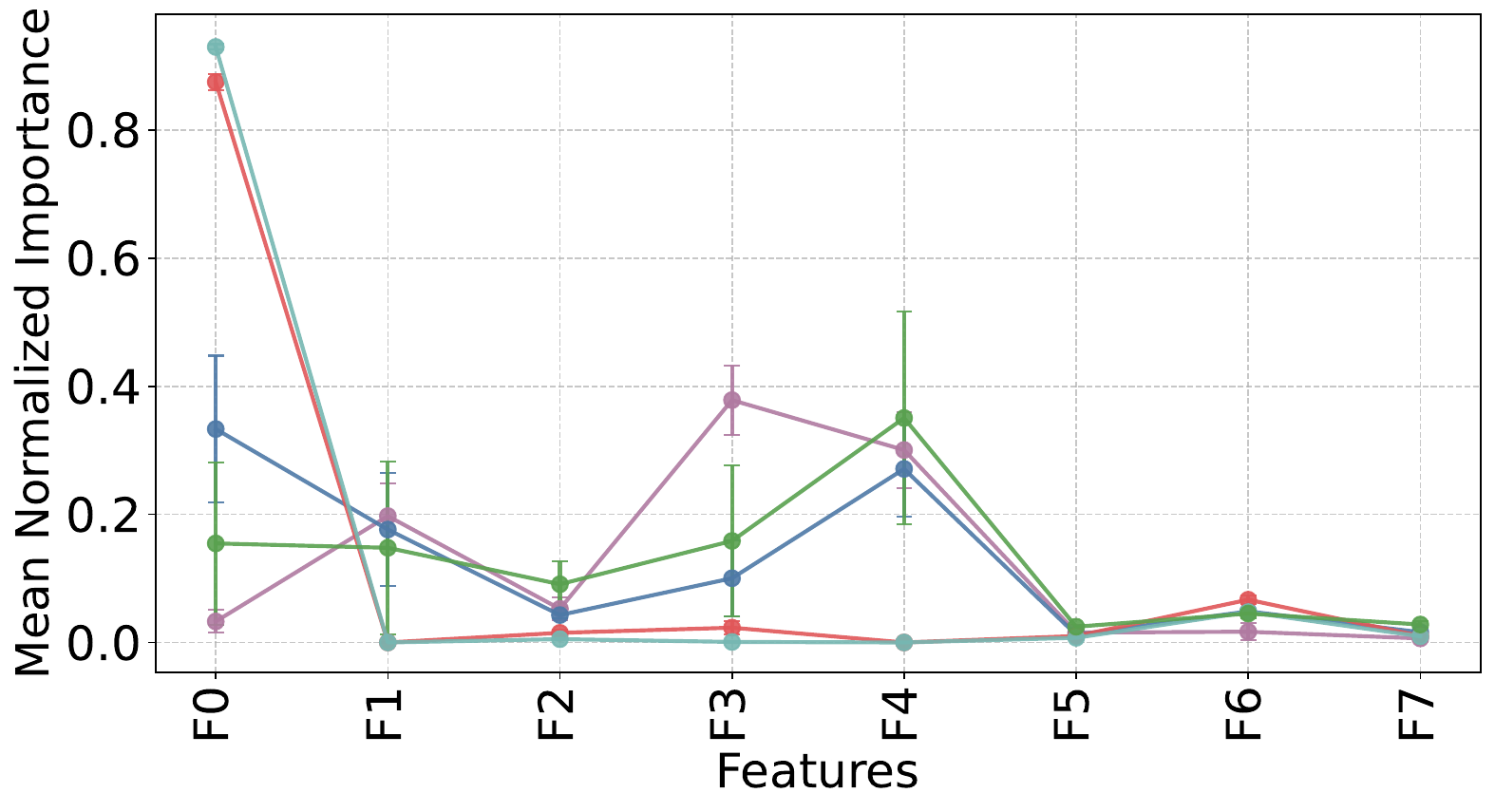}};
            \begin{scope}[x={(img.south east)}, y={(img.north west)}]
                \node[step=0.5, rotate=0] at (0.55, 0.9) {\color{black} \small EECooling Dataset};
            \end{scope}
        \end{tikzpicture}
    \end{subfigure}

    \vspace{0.3cm}

    \begin{subfigure}[b]{0.47\textwidth}
        \centering
        \begin{tikzpicture}
            \node[anchor=south west, inner sep=0] (img) at (0,0) {\includegraphics[width=\linewidth]{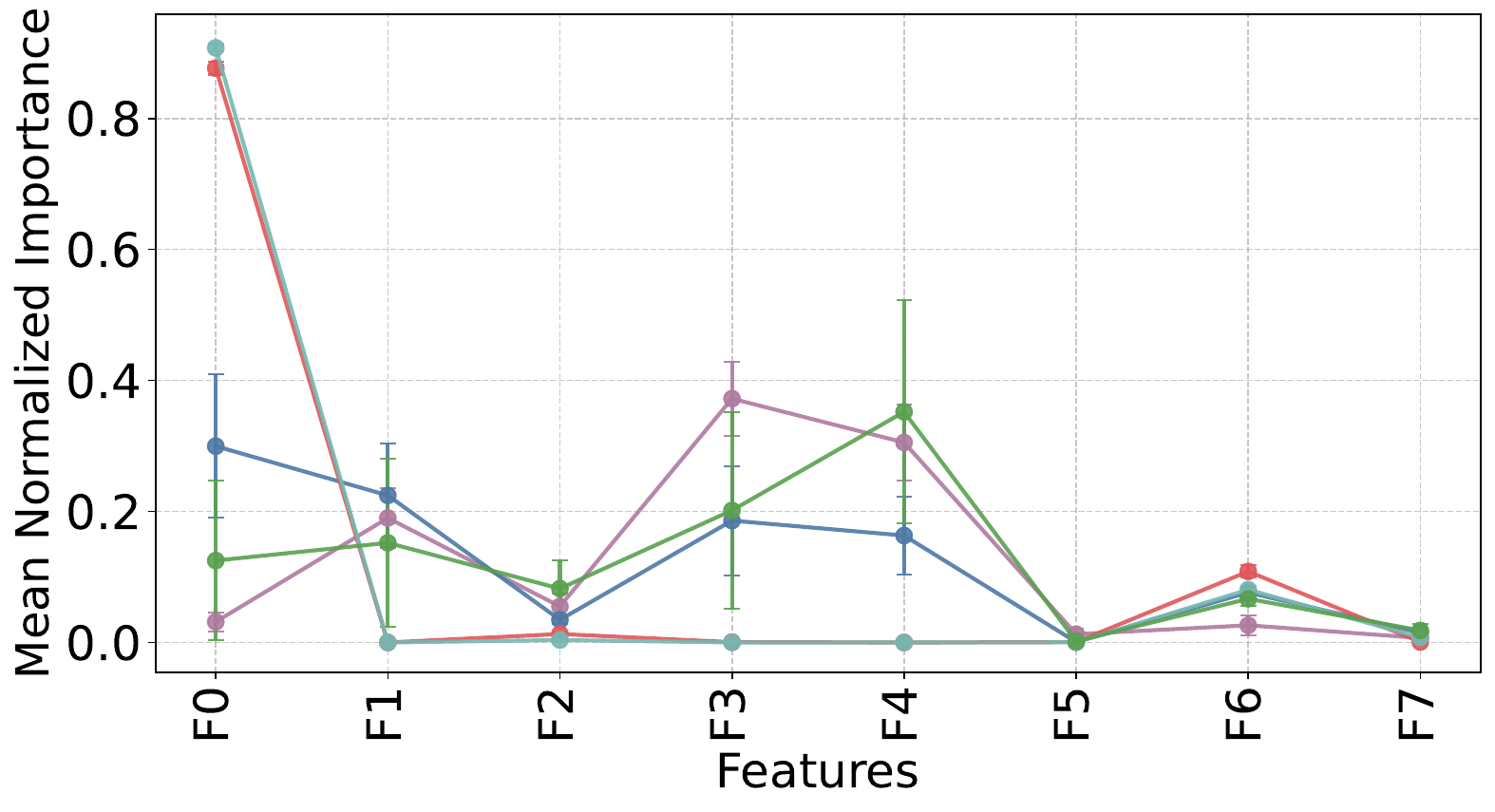}};
            \begin{scope}[x={(img.south east)}, y={(img.north west)}]
                \node[step=0.5, rotate=0] at (0.55, 0.9) {\color{black} \small EEHeating Dataset};
            \end{scope}
        \end{tikzpicture}
    \end{subfigure}
    \hfill
    \begin{subfigure}[b]{0.47\textwidth}
        \centering
        \begin{tikzpicture}
            \node[anchor=south west, inner sep=0] (img) at (0,0) {\includegraphics[width=\linewidth]{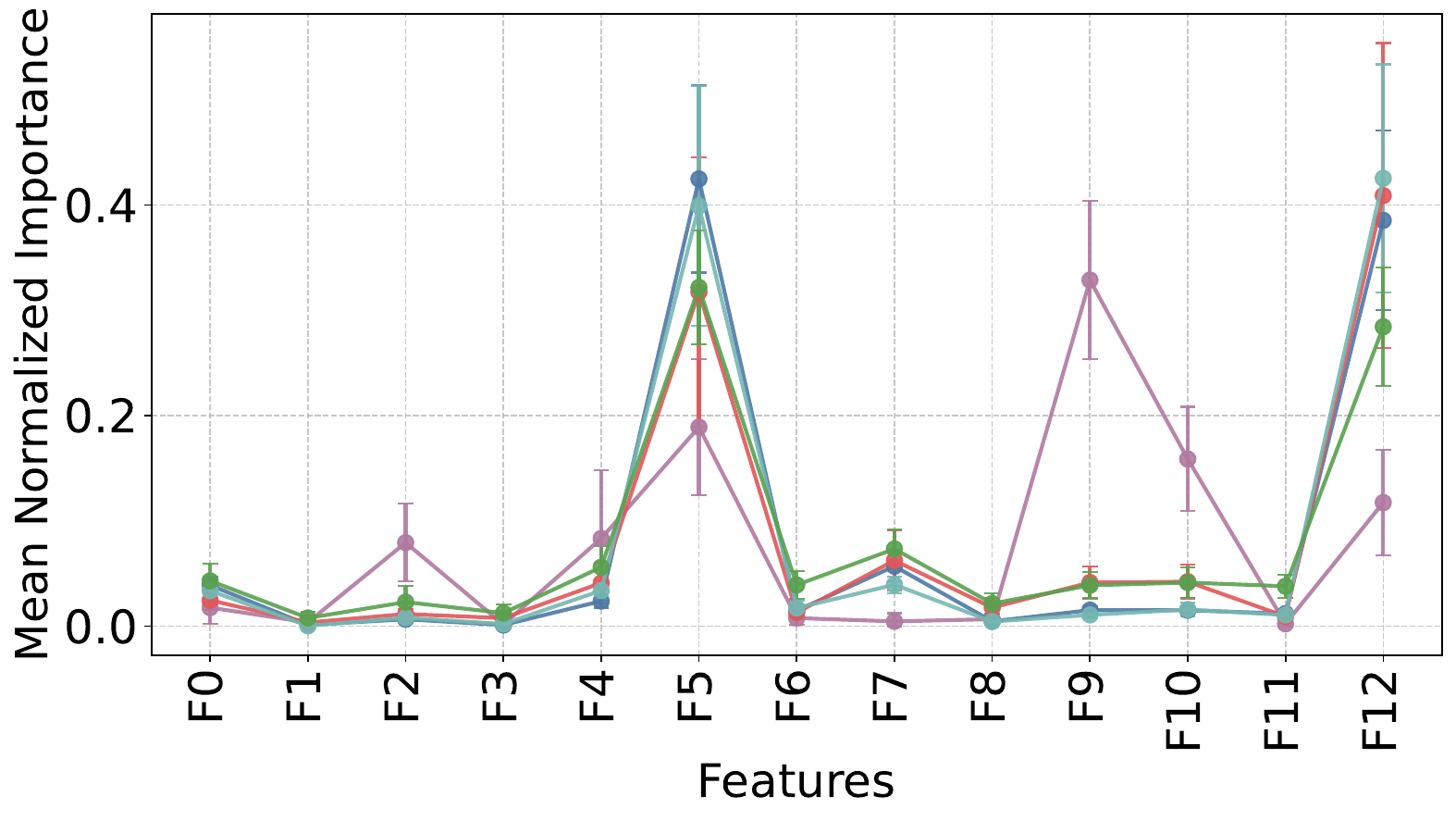}};
            \begin{scope}[x={(img.south east)}, y={(img.north west)}]
                \node[step=0.5, rotate=0] at (0.55, 0.9) {\color{black} \small Housing Dataset};
            \end{scope}
        \end{tikzpicture}
    \end{subfigure}

    \vspace{0.3cm}

    \begin{subfigure}[b]{0.47\textwidth}
        \centering
        \begin{tikzpicture}
            \node[anchor=south west, inner sep=0] (img) at (0,0) {\includegraphics[width=\linewidth]{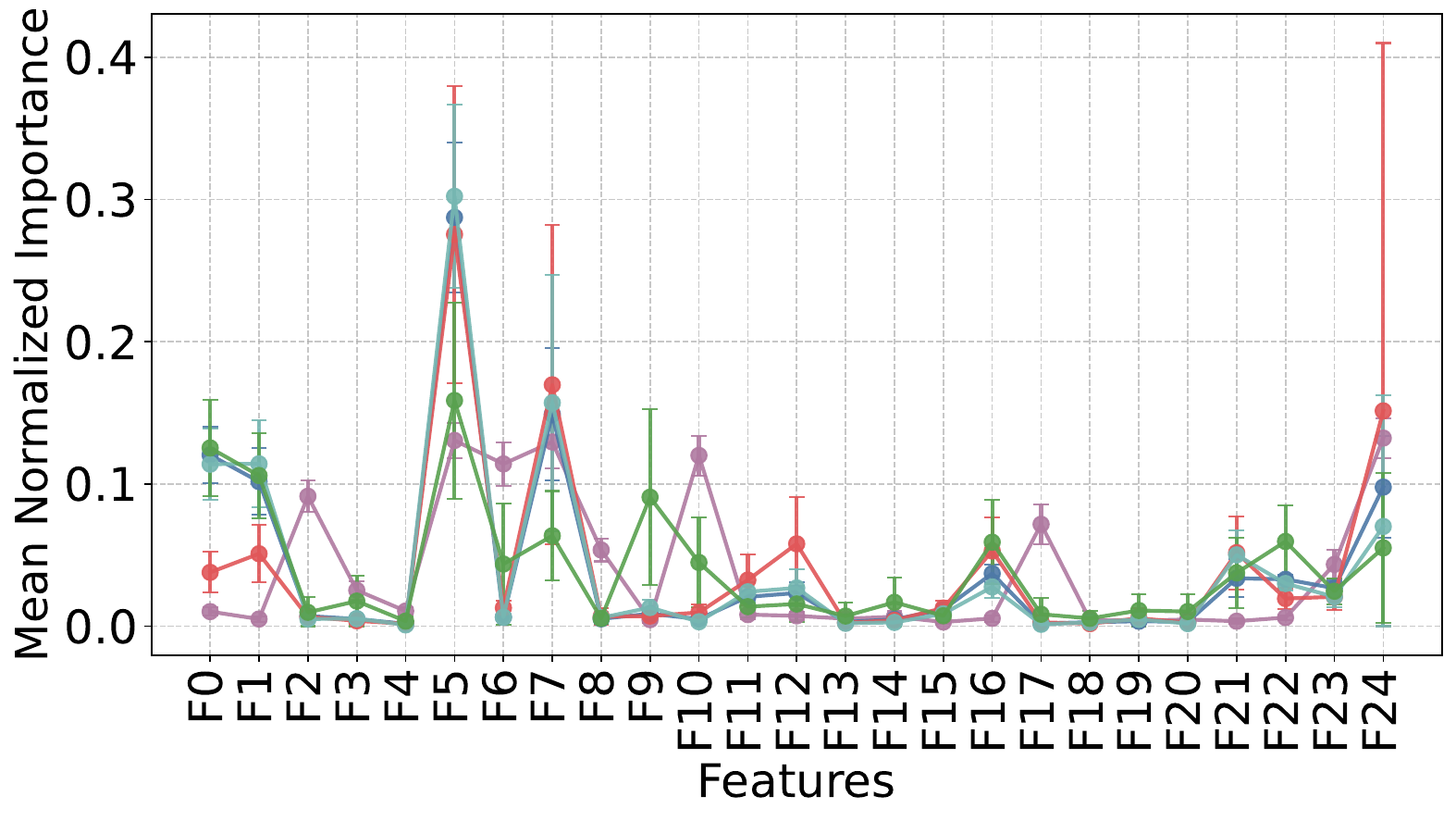}};
            \begin{scope}[x={(img.south east)}, y={(img.north west)}]
                \node[step=0.5, rotate=0] at (0.55, 0.9) {\color{black} \small Tower Dataset};
            \end{scope}
        \end{tikzpicture}
    \end{subfigure}
    \hfill
    \begin{subfigure}[b]{0.47\textwidth}
        \centering
        \begin{tikzpicture}
            \node[anchor=south west, inner sep=0] (img) at (0,0) {\includegraphics[width=\linewidth]{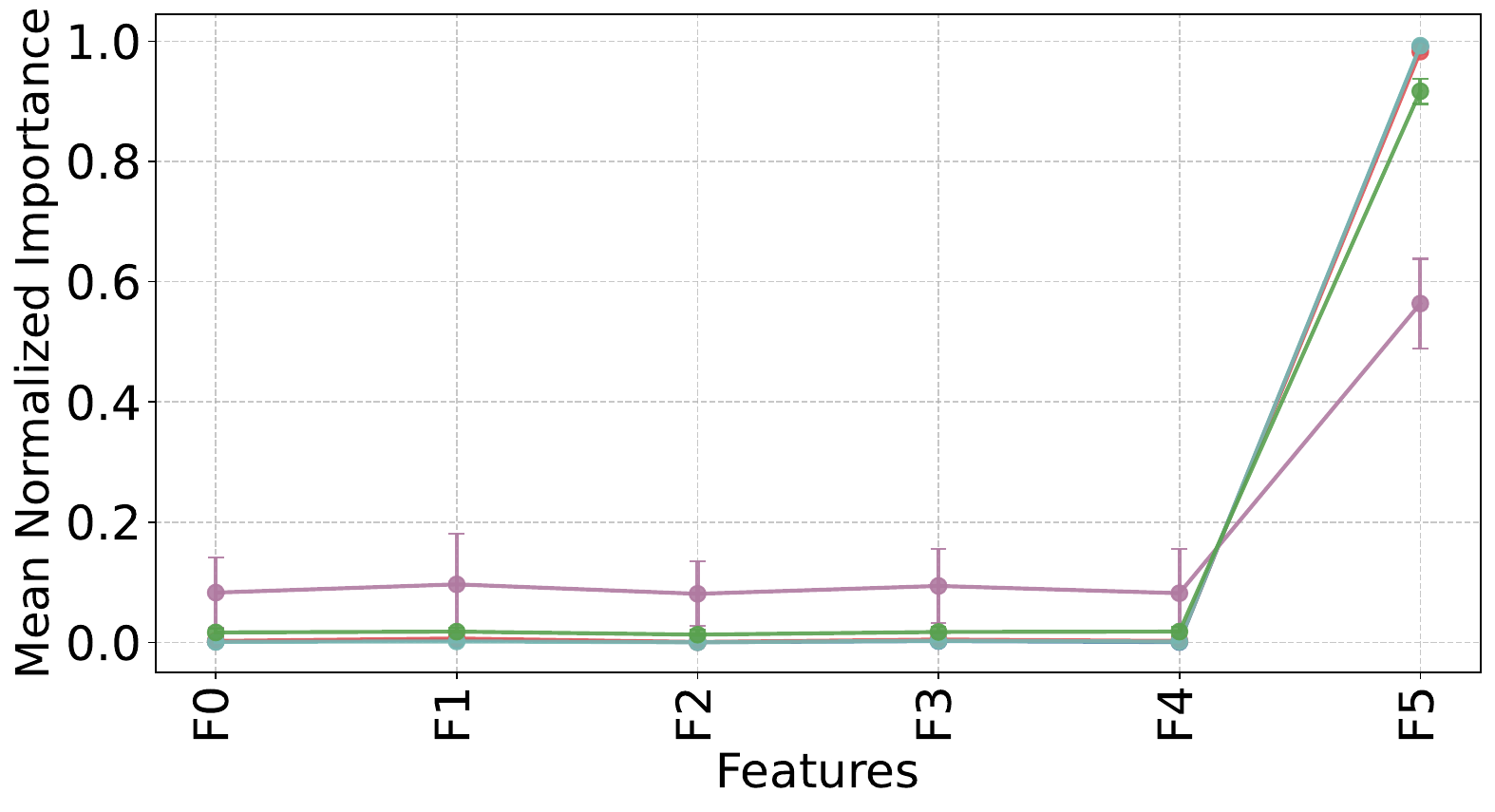}};
            \begin{scope}[x={(img.south east)}, y={(img.north west)}]
                \node[step=0.5, rotate=0] at (0.55, 0.9) {\color{black} \small Yacht Dataset};
            \end{scope}
        \end{tikzpicture}
    \end{subfigure}

    \vspace{0.3cm}

    \begin{minipage}{1\textwidth} 
        \centering 
        \scriptsize
        \renewcommand{\arraystretch}{1.5} 
        \begin{tabular}{cccccc} 
              \mylegend{RSB}{RSB} &
              \mylegend{RF}{Random Forest} &
              \mylegend{XGB}{XGBoost} &
              \mylegend{CatBoost}{CatBoost} &
              \mylegend{LGBM}{LightGBM} &
              \RefErrorBar \text{ SD}
        \end{tabular}
    \end{minipage}

    \caption{Comparison of feature importance between tree-based algorithms and RSB. The markers represent the mean importance.}
    \label{fig:feature_importance}
\end{figure*}

The results were compared with state-of-the-art tree-based algorithms, namely Random Forest (RF), XGBoost (XGB), LightGBM (LGBM), and CatBoost \citep{NEURIPS2022_0378c769}. Table \ref{table:model_params} details the configuration used; all methods were restricted to the same number of estimators ($N_{est}$) to ensure a fair comparison, without regularization and with early stopping disabled. Figure \ref{fig:feature_importance} shows the normalized mean importance for all variables in each dataset. The most notable difference is that RSB presents a peak in importance for variable F9 in the Housing dataset, which differs from the other algorithms. Furthermore, in the Yacht dataset, the tree ensembles attribute almost all the importance to F5, while RSB distributes the importance among the rest of the variables, keeping F5 as the most relevant one.

\begin{table}[t]
    \centering
    \caption{Hyperparameter settings for the comparative feature importance analysis.}
    \label{table:model_params}
    \begin{tabular}{ll}
        \toprule
        \textbf{Algorithm} & \textbf{Configuration} \\
        \midrule
        \textbf{RSB} & $m = 8$, $K=50$, $\eta=0.1$, $d_{max}=3$ \\
        RF & $N_{est}=100$, Library defaults \\
        XGB & $N_{est}=100$, Library defaults \\
        LGBM & $N_{est}=100$, Library defaults \\
        CatBoost & $N_{est}=100$, Library defaults \\
        \bottomrule
    \end{tabular}
\end{table}

\subsection{Ablation Study: Regularization and Early Stopping}\label{subsec:ablation_reg_es}

Elastic Net (ENet) and its special cases (OLS, Ridge, Lasso) were evaluated, each without ES, with the proposed angle-based ES, and with a conventional ES based on the Root Mean Squared Error (RMSE), for a total of twelve variants. The RMSE-based ES uses exactly the same mechanism as the angle-based one (Section~\ref{subsec:ages}), including the calibration window, dynamic tolerance (Eq.~\eqref{eq:tolerance}), convergence criterion (Eq.~\eqref{eq:convergence}), degradation slope (Eq.~\eqref{eq:slope}) and rollback, replacing $\theta_k$ with the validation RMSE at level $k$. Therefore, the two criteria differ only in the metric being monitored.

All variants share the same fixed hyperparameters: $m=8$, $K=200$, $\eta=0.1$, $d_{max}=3$, operators $\{+,-,\times,\mathrm{AQ},\sin\}$, standardization of the semantic matrix, and, for the penalized cases, $\lambda=0.02$ with $\rho=0$ for Ridge, $\rho=1$ for Lasso, and $\rho=0.5$ for ENet. The ES variants use $W=15$, $\gamma=0.1$, $P_c=10$, $L=15$, $\beta_{tol}=0.1$, and $P_d=10$. Each variant was run 30 times on each dataset of Table~\ref{tab:datasets}, recording the test $R^2$ (clipped at 0), the number of nodes, and the training time. In each run, all variants share the same random split: without ES, 80\% of the data is used for training and 20\% for testing; with ES, the training portion is further divided into 60\% for training and 20\% for validation (60/20/20), and the test set is the same. In each case, $n$ in $M = m\sqrt{n}$ is the size of the 80\% partition, so the number of trees per level is identical across variants. Table~\ref{tab:median_variant_comparison} reports, for each variant, the median across the six datasets of the per-dataset median over the 30 runs.

To assess significance, the Nadeau--Bengio corrected resampled $t$-test with Holm correction \citep{nadeaubengio2003, holm1979} was applied, separately for each dataset and metric ($\alpha=0.05$), pooling each family's comparisons in one correction batch. Three comparison families were made: (i) all pairs among the four variants without ES (each variant against 3 rivals on 6 datasets), (ii) each variant against its own angle-based and RMSE-based ES versions, and (iii) angle-based against RMSE-based ES for each regularization. When the Holm-adjusted $p$-value is below $\alpha$, the comparison is significant and one variant wins and the other loses (higher $R^2$, fewer nodes, or lower time); otherwise it is non-significant (NS). Table~\ref{tab:nadeau_holm_variant_comparison} summarizes the results. The first block shows, for each variant without ES, the number of comparisons it wins, loses, or leaves NS against the other three. The second and third blocks show, for each regularization, the number of datasets in which each ES criterion is better, worse, or NS than no ES, and the last block compares the two criteria.

Among the variants without ES, Ridge never loses in $R^2$ (9 wins, 9 NS) and is the fastest (17 wins, 1 loss), but it barely reduces the model size (5 wins, 13 losses). Lasso produces the smallest models (it wins all 18 comparisons in nodes) with an $R^2$ comparable to ENet (4 wins, 4 losses, and 10 NS for each), and is the second fastest. ENet is intermediate in $R^2$ and nodes (12 wins and 6 losses in nodes) and the slowest (1 win, 16 losses). OLS never wins in $R^2$.

Regarding the angle-based ES, it significantly reduces both nodes and training time on all six datasets for OLS, Lasso, and ENet, and on five of six for Ridge (one NS). It does not significantly improve $R^2$ for any penalized variant: it is NS on four of six datasets for Ridge and on two of six for Lasso and ENet, and it decreases $R^2$ on the rest (2 of 6 for Ridge, 4 of 6 for Lasso and ENet). The only improvement appears for OLS (3 of 6), where the angle-based ES prevents the collapse. In terms of medians, the angle-based ES reduces the nodes of the penalized variants by 54--65\% at a cost of 0.003 in $R^2$ for Ridge and 0.017 for Lasso and ENet. The RMSE-based ES also reduces nodes and training time on all six datasets for every variant, but it decreases $R^2$ on five of six datasets for each penalized variant.

Overall, among the variants without ES, Ridge offers the best trade-off between $R^2$ and training time, while Lasso offers the best trade-off between model size and time; ENet does not stand out on any of the three metrics. Ridge with angle-based ES attains a median $R^2$ of 0.970 (0.003 below Ridge without ES), about 65\% fewer nodes than Ridge without ES, and the lowest training time among the angle-based variants (0.33~s), at the cost of larger models than Lasso with angle-based ES. Lasso with angle-based ES yields the smallest models (18,900 nodes) at the cost of 0.022 in $R^2$ relative to Ridge with angle-based ES. Ridge with angle-based ES is therefore adopted in the benchmarks of Section~\ref{benchmarking}.

Finally, the last block of Table~\ref{tab:nadeau_holm_variant_comparison} compares the two ES criteria directly. For OLS, no significant difference is found on any dataset or metric. The angle is significantly better in $R^2$ on two of six datasets for Ridge and on three of six for Lasso and ENet, and it is never significantly worse. In exchange, RMSE-based ES produces smaller models, being significantly better in nodes on four of six datasets for Ridge and on three of six for Lasso and ENet, and in time on two of six for Ridge, four of six for Lasso, and five of six for ENet. The angle thus favors accuracy, while the RMSE favors compactness and speed.

\begin{table}[t]
    \centering
    \caption{Median test $R^2$, number of nodes, and training time for OLS, Ridge, Lasso, and ENet without ES; with angle-based ES; and with RMSE-based ES. Each value is the median across the six datasets of the per-dataset median over 30 runs. Bold indicates the best value per column.}
    \label{tab:median_variant_comparison}
    \resizebox{\columnwidth}{!}{%
    \begin{tabular}{lccc}
        \toprule
        \textbf{Variant} & \textbf{\boldmath$R^2$} & \textbf{Nodes ($\times10^3$)} & \textbf{Time (s)} \\ \midrule
        OLS & 0.429 & 157.1 & 3.74 \\ 
        Ridge & \textbf{0.973} & 151.8 & 0.69 \\ 
        Lasso & 0.965 & 41.9 & 1.20 \\ 
        ENet & 0.953 & 61.3 & 5.82 \\ 
        OLS+ES (angle) & 0.902 & 35.7 & 0.92 \\ 
        Ridge+ES (angle) & 0.970 & 53.6 & 0.33 \\ 
        Lasso+ES (angle) & 0.948 & 18.9 & 0.63 \\ 
        ENet+ES (angle) & 0.936 & 28.4 & 4.68 \\ 
        OLS+ES (RMSE) & 0.895 & 25.6 & 0.74 \\ 
        Ridge+ES (RMSE) & 0.963 & 29.0 & \textbf{0.20} \\ 
        Lasso+ES (RMSE) & 0.941 & \textbf{12.5} & 0.48 \\ 
        ENet+ES (RMSE) & 0.926 & 17.6 & 3.95 \\ \bottomrule
    \end{tabular}%
    }
\end{table}

\begin{table*}[t]
    \centering
    \caption{Summary of pairwise comparisons (Nadeau--Bengio corrected $t$-test with Holm correction per dataset and metric, $\alpha=0.05$; 30 runs per dataset). Reg.: each method without ES vs.\ the other three (wins/losses/ns over 18 comparisons). Angle-based ES / RMSE-based ES vs.\ no ES: datasets where each ES criterion is better/worse/ns than no ES (higher $R^2$, fewer nodes, or lower time). Angle- vs.\ RMSE-based ES: datasets where angle-based ES is better/worse/ns than RMSE-based ES. Bold indicates the best result per metric within the Reg. block.}
    \label{tab:nadeau_holm_variant_comparison}
    \resizebox{\textwidth}{!}{%
    \begin{tabular}{lcccccccccccc}
        \toprule
        & \multicolumn{3}{c}{\textbf{Reg.\ (wins/losses/ns)}} &
        \multicolumn{3}{c}{\textbf{Angle-based ES vs. no ES}} &
        \multicolumn{3}{c}{\textbf{RMSE-based ES vs. no ES}} &
        \multicolumn{3}{c}{\textbf{Angle-based ES vs. RMSE-based ES}} \\ \cmidrule(lr){2-4}\cmidrule(lr){5-7}\cmidrule(lr){8-10}\cmidrule(lr){11-13}
        Method & $R^2$ & Nodes & Time &
        $R^2$ & Nodes & Time &
        $R^2$ & Nodes & Time &
        $R^2$ & Nodes & Time \\ \midrule
        OLS & 0/9/9 & 1/17/0 & 7/11/0 & 3/0/3 & 6/0/0 & 6/0/0 & 3/0/3 & 6/0/0 & 6/0/0 & 0/0/6 & 0/0/6 & 0/0/6 \\ 
        Ridge & \textbf{9/0/9} & 5/13/0 & \textbf{17/1/0} & 0/2/4 & 5/0/1 & 5/0/1 & 0/5/1 & 6/0/0 & 6/0/0 & 2/0/4 & 0/4/2 & 0/2/4 \\ 
        Lasso & 4/4/10 & \textbf{18/0/0} & 10/7/1 & 0/4/2 & 6/0/0 & 6/0/0 & 0/5/1 & 6/0/0 & 6/0/0 & 3/0/3 & 0/3/3 & 0/4/2 \\ 
        ENet & 4/4/10 & 12/6/0 & 1/16/1 & 0/4/2 & 6/0/0 & 6/0/0 & 0/5/1 & 6/0/0 & 6/0/0 & 3/0/3 & 0/3/3 & 0/5/1 \\ \bottomrule
    \end{tabular}%
    }
\end{table*}

\section{Benchmarking}\label{benchmarking}

This section evaluates the predictive performance of RSB on two standard benchmarks: SRBench \citep{lacava2021contemporary} for regression, and TALENT \citep{JMLR:v26:25-0512} for binary and multiclass classification. In both cases, the results of the baseline methods were taken from the official repositories of each benchmark, so that the comparison is made under the protocols originally defined by their authors. Section~\ref{srbench} reports the results for regression in terms of accuracy, training time, model size, statistical significance and Pareto front, while Section~\ref{talent} presents the results for classification.

\subsection{SRBench}\label{srbench}

To evaluate RSB in regression, the SRBench protocol established in \citet{lacava2021contemporary} was followed, which includes 122 black-box problems (62 Synthetic and 60 Real-World). The results of the baseline methods were obtained from the SRBench repository\footnote{Repository available at: \url{https://github.com/cavalab/srbench}}, executed on a cluster with Intel Xeon E5-2690 v4 (2.60 GHz) nodes and 250 GB of RAM. Meanwhile, RSB was executed on a local workstation with two Intel Xeon E5-2603 v4 CPUs at 1.70 GHz (12 cores), 40 GB of RAM, a 2 TB SSD, and Ubuntu 20.04 LTS. A single core was used per execution and maximum number of evaluations were respected. No GPU acceleration was employed.

RSB was evaluated in only 6 configurations, as SRBench restricts the search to 6 combinations for symbolic regression methods due to the high computational cost inherent to these evolutionary methods \citep{lacava2021contemporary}. The hyperparameter configurations used are shown in Table \ref{table:hyperparams_srbench}. Ridge regularization with angle-based ES was chosen based on the ablation in Section~\ref{subsec:ablation_reg_es}. The depths $d_{max}\in\{1,3\}$ follow from the fANOVA in Section~\ref{experimental}, where $d_{max}$ explained little of the variance in $R^2$ relative to $K$ and $\eta$. Standardization was applied as the transformation of the semantic matrix.

\begin{figure*}[!tp]
    \centering
    \begin{subfigure}[b]{0.425\textwidth}
        \centering
        \includegraphics[width=\linewidth]{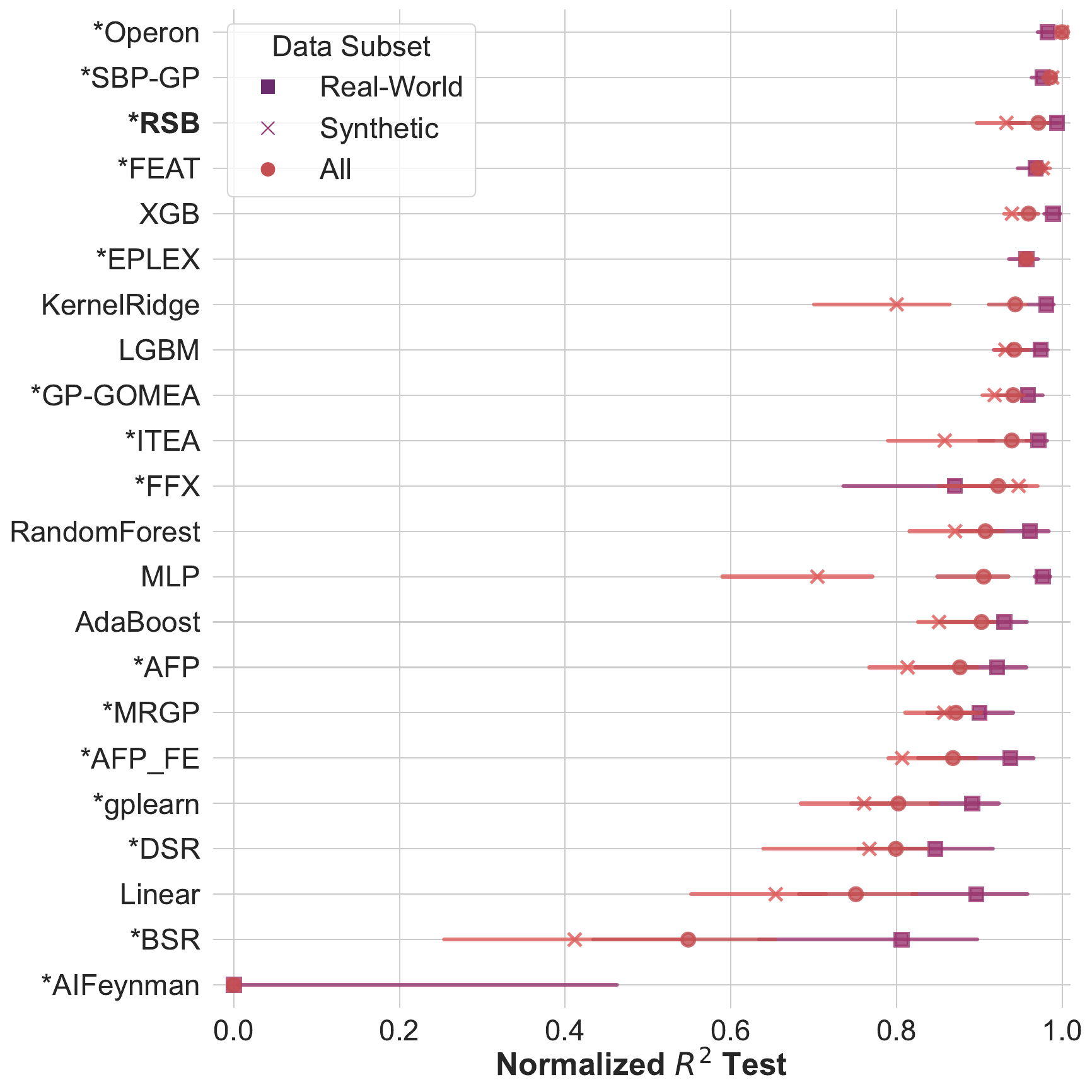}
        \caption{}
    \end{subfigure}
    \hfill
    \begin{subfigure}[b]{0.28\textwidth}
        \centering
        \includegraphics[width=\linewidth]{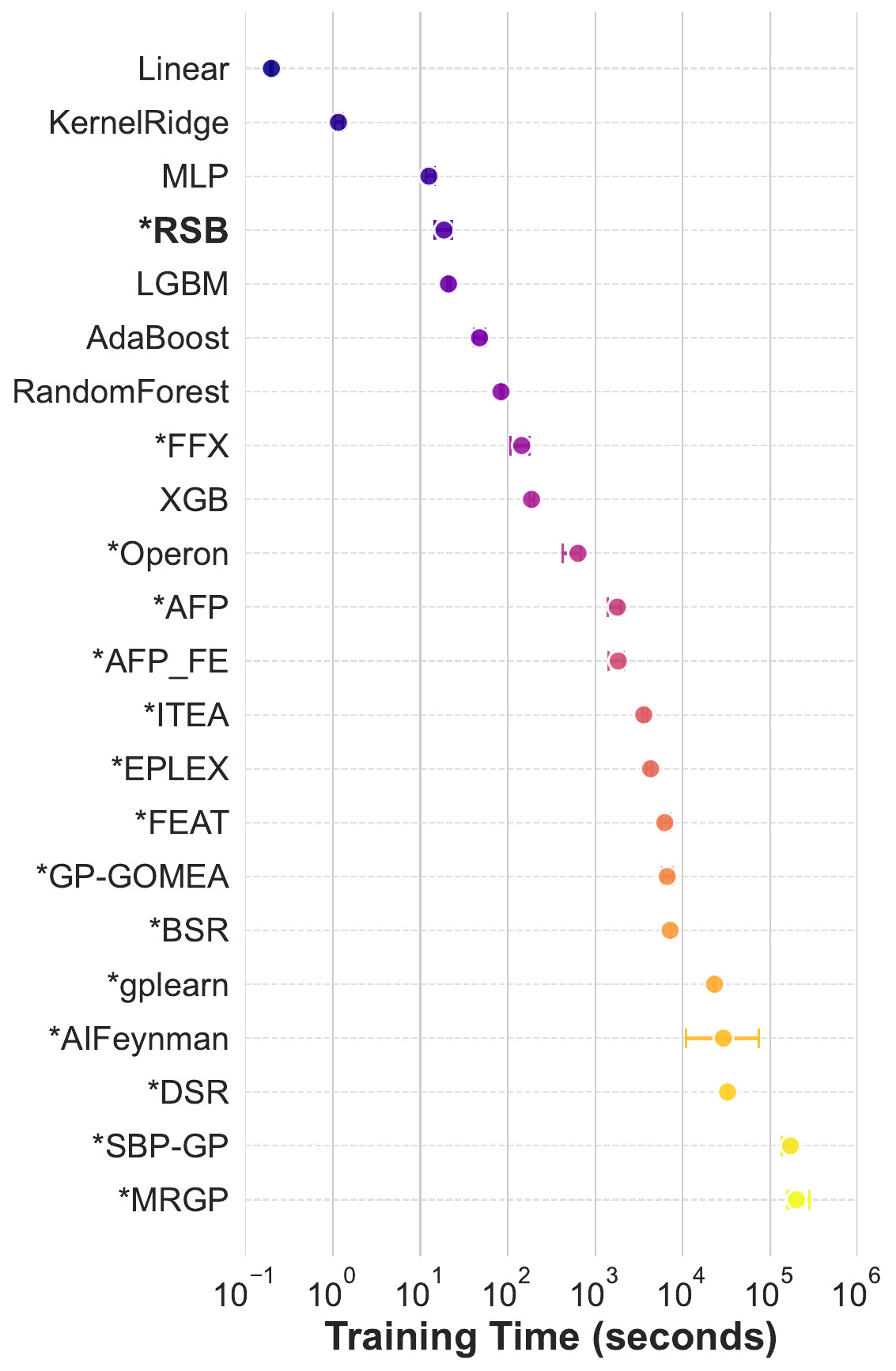}
        \caption{}
    \end{subfigure}
    \hfill
    \begin{subfigure}[b]{0.28\textwidth}
        \centering
        \includegraphics[width=\linewidth]{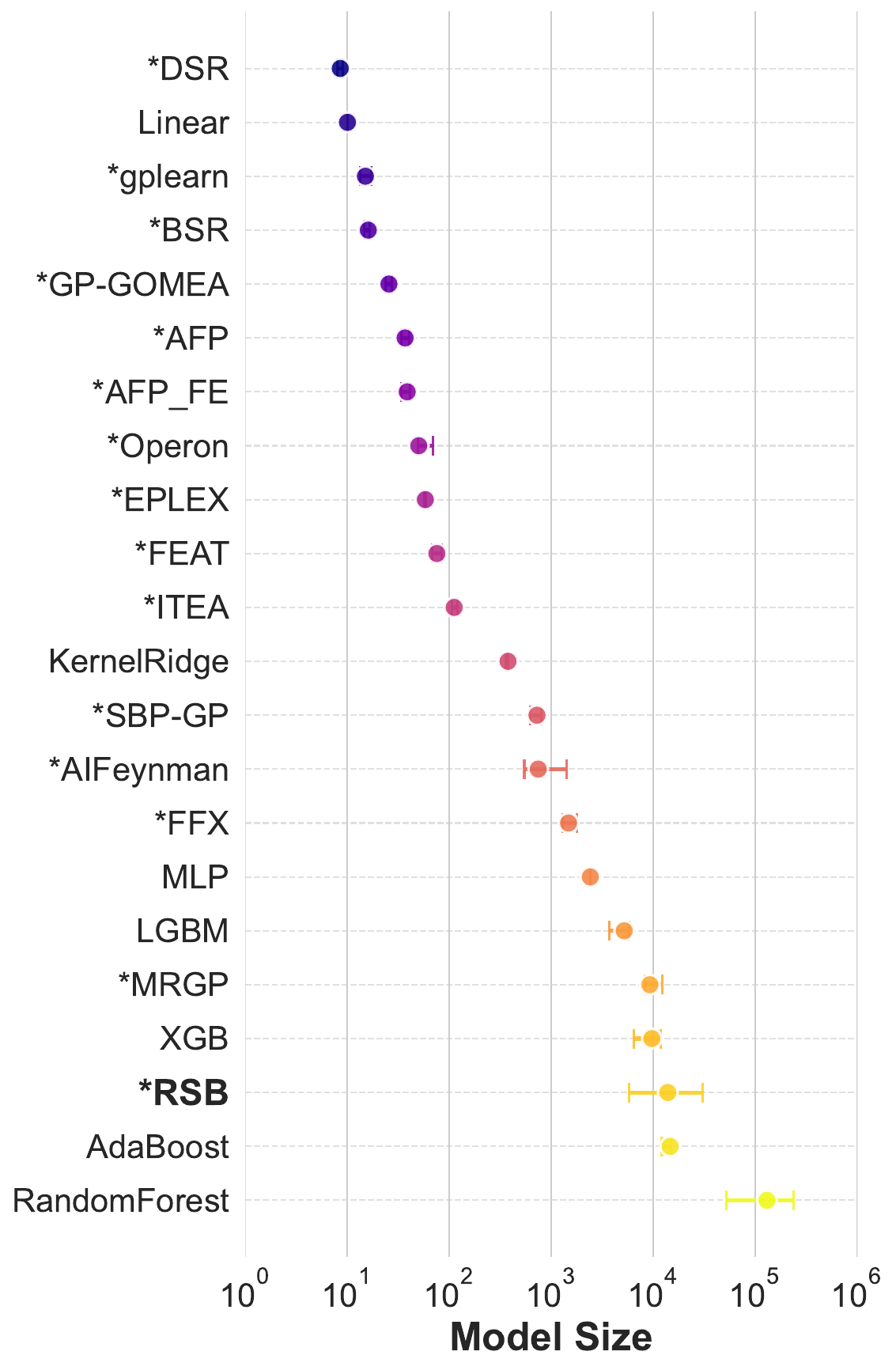}
        \caption{}
    \end{subfigure}

    \caption{Comparison of the RSB algorithm against baseline methods in SRBench: (a) normalized $R^2$ test score, (b) training time, and (c) model size.}
    \label{fig:metrics_srbench}
\end{figure*}

Figure \ref{fig:metrics_srbench} shows the results obtained in $R^2$, training time, and model size.

\begin{table}[t]
    \centering
    \caption{Hyperparameter search space used for the evaluation in SRBench.}
    \label{table:hyperparams_srbench}
    \begin{tabular}{lc}
        \toprule
        \textbf{Hyperparameter} & \textbf{Evaluated Values} \\
        \midrule
        $m$ & $\{1, 5, 25\}$ \\
        $K$ & $50$ \\
        $\eta$ & $0.1$ \\
        $d_{max}$ & $\{1, 3\}$ \\
        $\lambda, \rho$ & $0.02$, $0$ \\
        Operators & $\{+, -, \times, \text{AQ}, \sin\}$ \\
        $W, \gamma, P_{c}, L, \beta_{tol}, P_{d}$ & $15,\; 0.1,\; 10,\; 15,\; 0.1,\; 10$ \\
        \bottomrule
    \end{tabular}
\end{table}

\subsubsection{Robustness and Global Convergence}

\begin{figure*}[!tbp]
    \centering
    \includegraphics[width=0.7\textwidth]{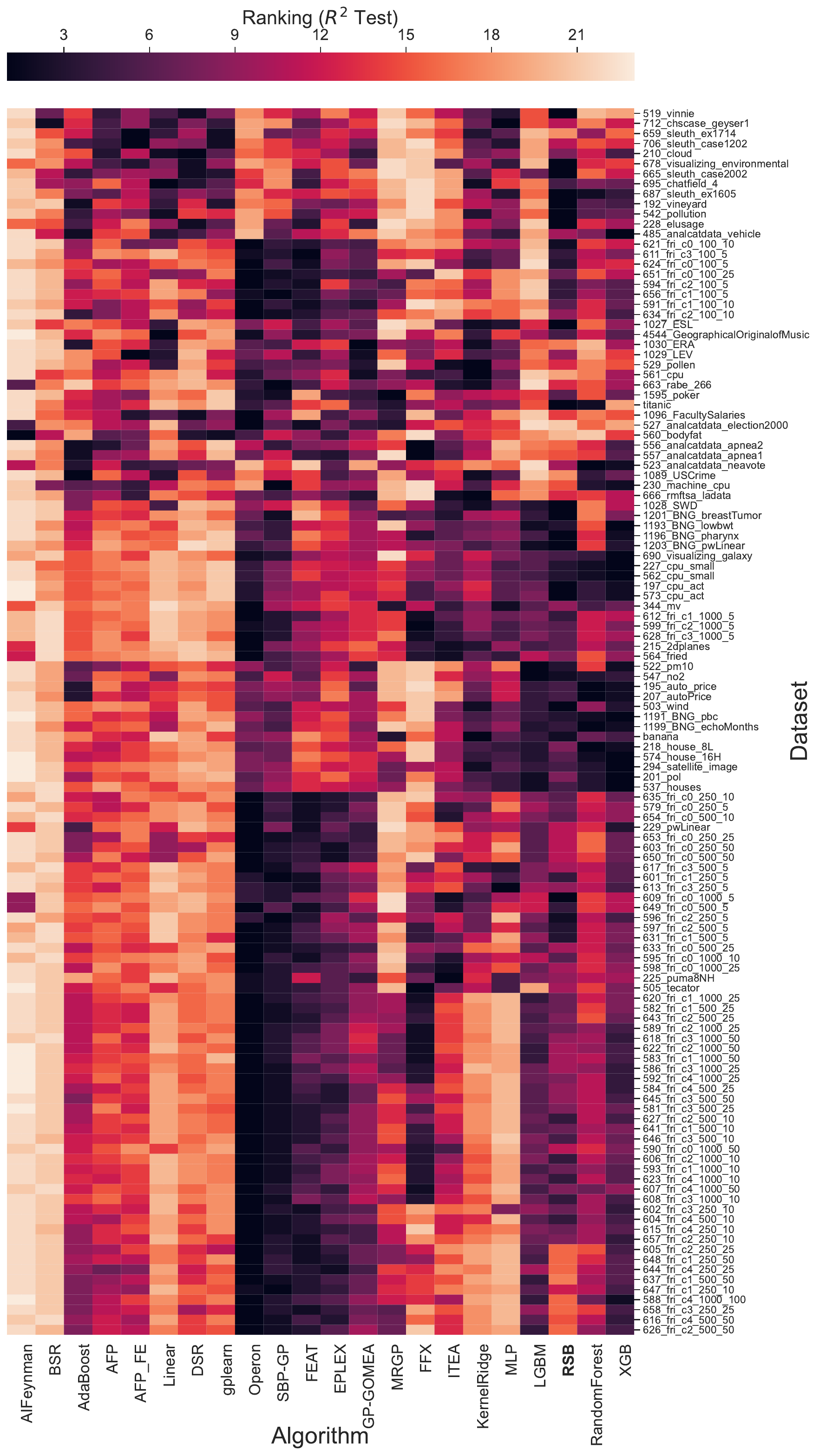}
    \caption{Rankings of methods by R$^2$ test score on the black-box problems (lower is better). Results are bi-clustered by algorithm and dataset. Darker cells indicate that a method performs well on that dataset relative to its competitors.}
    \label{fig:heatmap_rankings}
\end{figure*}

Regarding $R^2$, RSB ranks 3\textsuperscript{rd}, outperforming methods such as XGBoost and FEAT in the overall result (red circle), while achieving first place on the real-world datasets (purple square). To visualize the relative performance of the methods across the different datasets, Figure \ref{fig:heatmap_rankings} presents a heatmap of the $R^2$ rankings. It can be seen that RSB achieves the best overall performance on the real-world datasets, with the majority of dark-colored cells. On the other hand, regarding time efficiency, it ranked 4\textsuperscript{th}. It was only outperformed by classical linear models (Linear, KernelRidge) and simple networks (MLP). Finally, when examining model size, it is observed that RSB tends to generate large expressions due to the additive nature of the ensemble, ranking third to last, only above AdaBoost and Random Forest.

\subsubsection{Statistical Significance}

\begin{figure*}[!tp]
    \centering
    \begin{subfigure}[b]{0.325\textwidth}
        \centering
        \includegraphics[width=\linewidth]{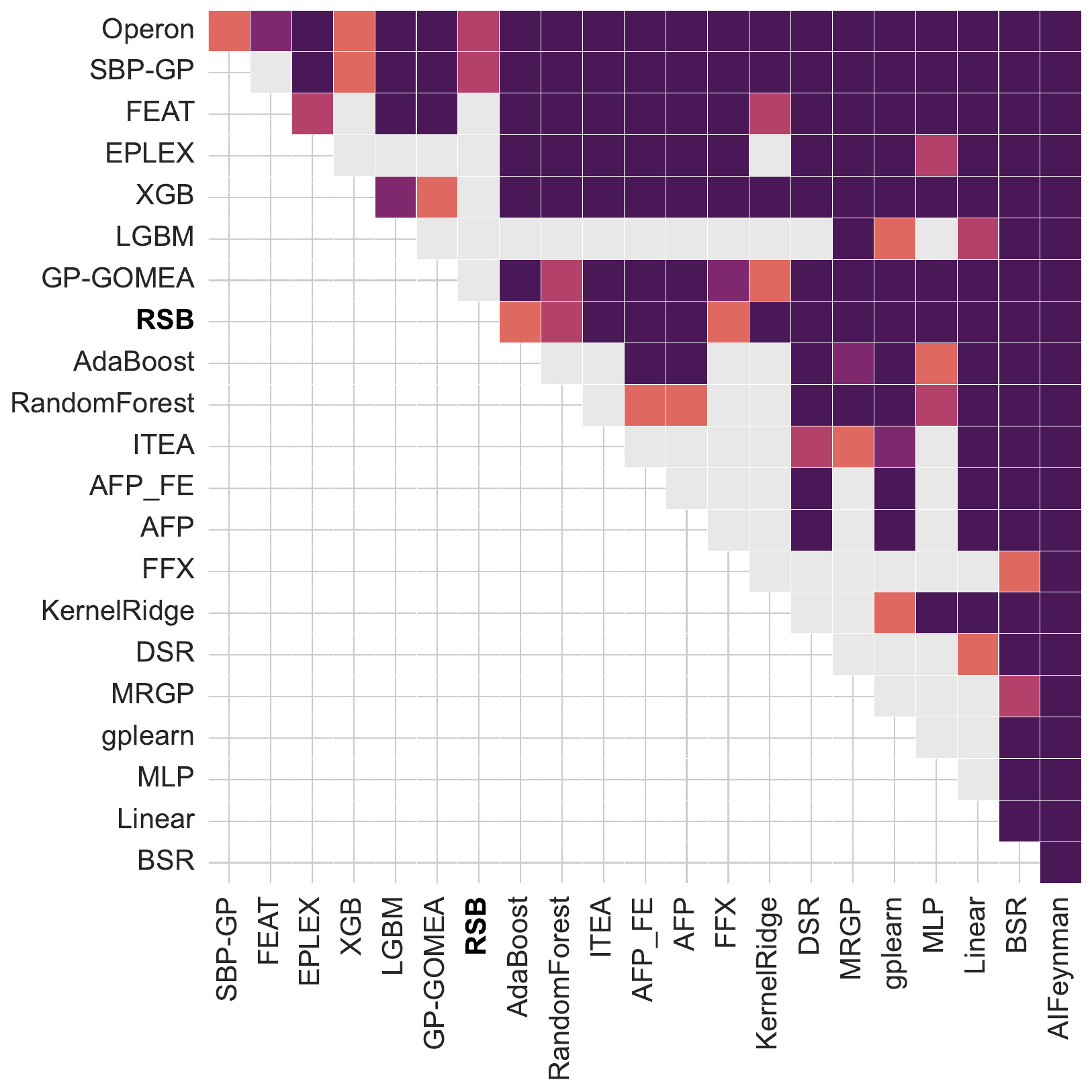}
        \caption{$R^2$}
    \end{subfigure}
    \hfill
    \begin{subfigure}[b]{0.325\textwidth}
        \centering
        \includegraphics[width=\linewidth]{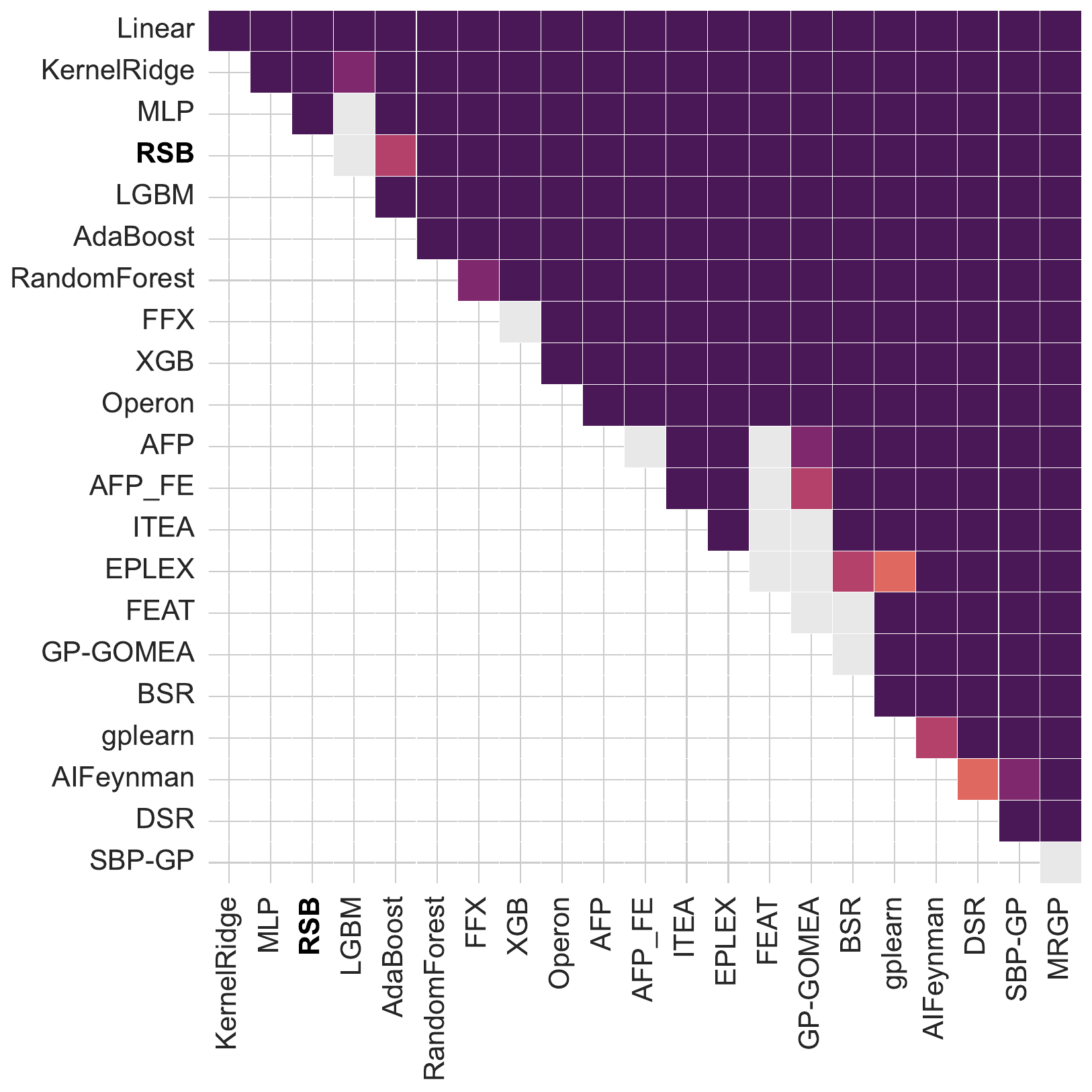}
        \caption{Training Time}
    \end{subfigure}
    \hfill
    \begin{subfigure}[b]{0.32\textwidth}
        \centering
        \includegraphics[width=\linewidth]{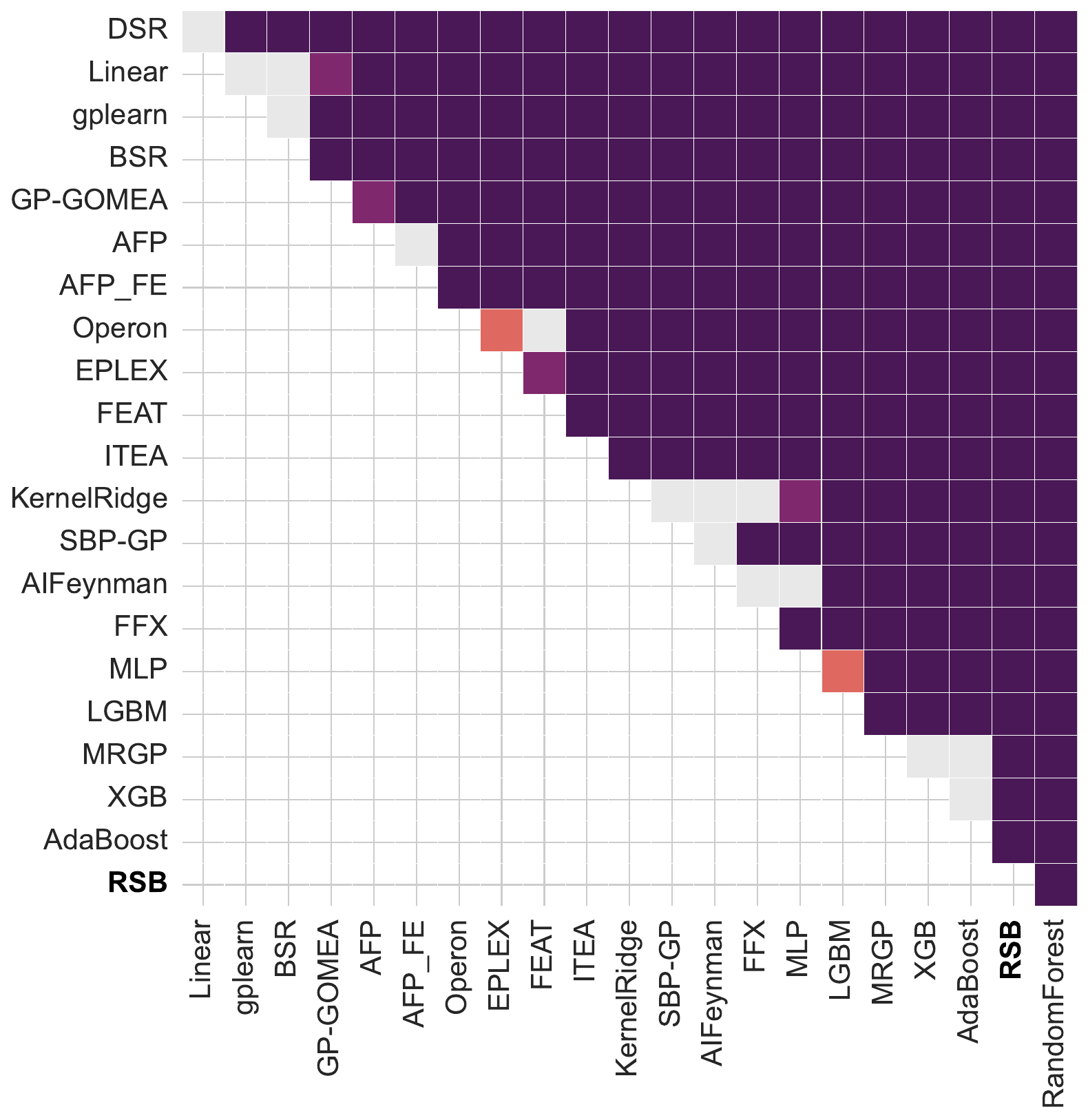}
        \caption{Model Size}
    \end{subfigure}

    \vspace{0.3cm}

    \begin{minipage}{1\textwidth} 
        \centering 
        \footnotesize
        \begin{tabular}{lllll} 
              \mylegend{wil_bon_3}{$p < 10^{-3} \cdot \alpha$} &
              \mylegend{wil_bon_2}{$p < 10^{-2} \cdot \alpha$} &
              \mylegend{wil_bon_1}{$p < 10^{-1} \cdot \alpha$} &
              \mylegend{wil_bon_0}{$p < \alpha$} &
              \mylegend{wil_bon_no_sign}{No significance}
        \end{tabular}
    \end{minipage}

    \caption{Statistical significance heatmaps using the pairwise Wilcoxon signed-rank test with Bonferroni correction ($\alpha=0.05$).}
    \label{fig:wilcoxon_r2_time_size}
\end{figure*}

To determine if there are statistically significant differences between the performance of RSB and the baseline algorithms across the multiple evaluated datasets, the non-parametric Wilcoxon signed-rank test with Bonferroni correction was applied \citep{demsar2006statistical}. Figure \ref{fig:wilcoxon_r2_time_size} presents the results of these comparisons for the metrics of accuracy ($R^2$), computational efficiency (Training Time), and model complexity (Model Size), respectively. Regarding $R^2$, it is observed that RSB achieves a performance that shows no statistically significant differences compared to black-box models such as XGBoost and LGBM, nor compared to symbolic methods such as FEAT, EPLEX, and GP-GOMEA. Regarding training time, RSB presents a statistically significant difference compared to almost all methods, except LGBM. Finally, regarding model complexity, the only algorithm with which there is no significant difference is XGB.

\subsubsection{Pareto Front Analysis}

\begin{figure}[h]
    \centering
    \includegraphics[width=\columnwidth]{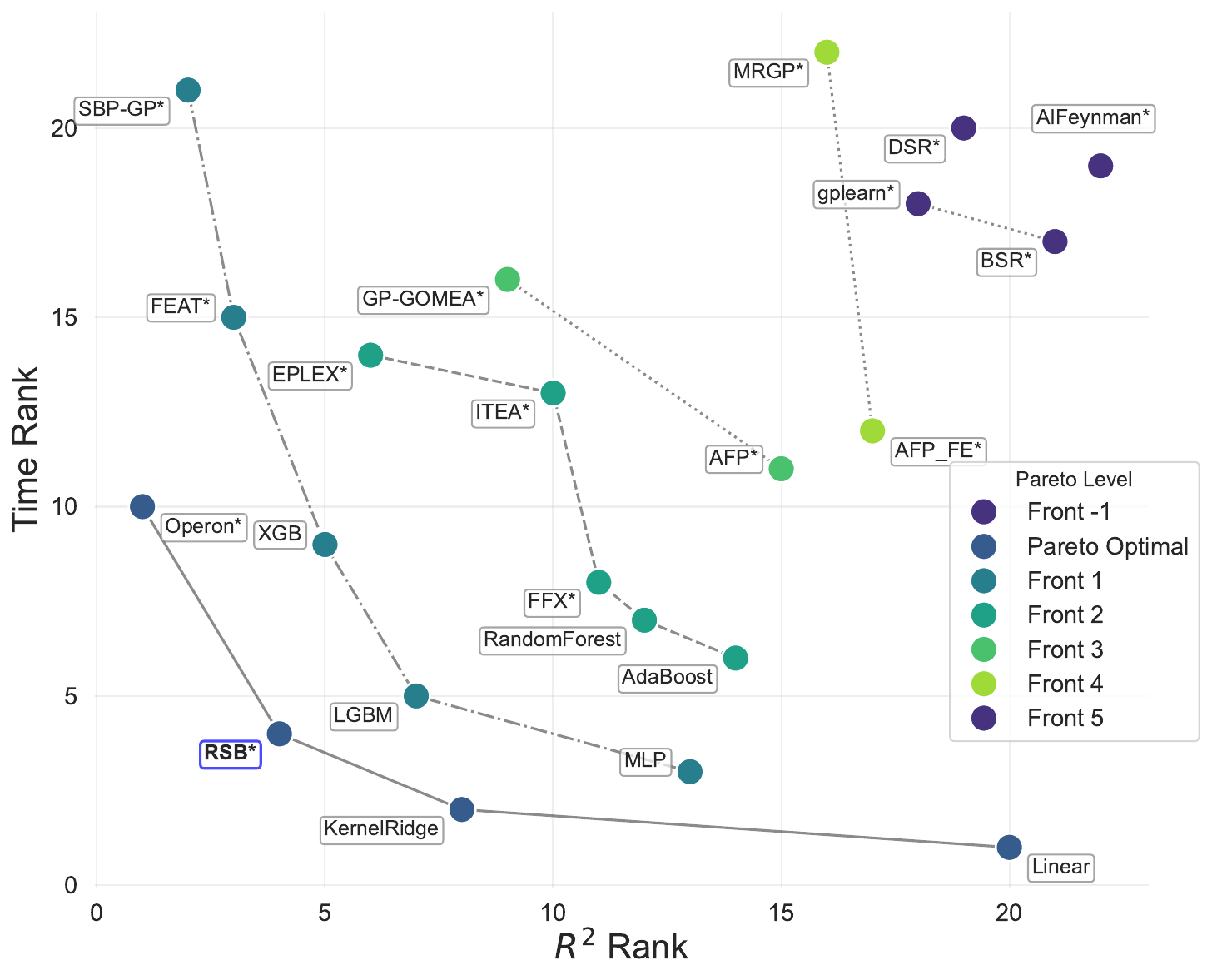}
    \caption{Pareto front between predictive accuracy ($R^2$ Ranking) and computational cost (Training Time Ranking).}
    \label{fig:pareto_time_r2}
\end{figure}

Figure \ref{fig:pareto_time_r2} shows the Pareto chart of Time Rank vs. $R^{2}$ Rank, where the point $(1,1)$ represents the ideal point of minimum computational complexity and maximum predictive performance. It is observed that RSB lies on the Pareto-optimal front, alongside Operon, KernelRidge, and Linear. It is worth noting that RSB is situated at the elbow of the optimal frontier, demonstrating that it is the model with the best overall trade-off; it offers a predictive accuracy comparable to Operon, but with greater computational efficiency, approaching the speed of linear models.

\subsection{TALENT}\label{talent}

To evaluate RSB in classification, the TALENT (Tabular Analytics and Learning Toolbox) benchmark \citep{JMLR:v26:25-0512} was used, which includes a diverse set of 120 binary classification problems and 80 multiclass classification problems. The results of the 31 baseline algorithms were obtained from the official TALENT repository\footnote{Repository available at: \url{https://github.com/LAMDA-Tabular/TALENT}}, executed under their conditions (the reference hardware is not specified in the repository). RSB was evaluated using the same hardware described in Section \ref{comp_cost}.

For the extension to multiclass classification with $C$ classes, the algorithm was adapted by treating the problem as a vector boosting task. Instead of a single estimator $F(\mathbf{x})$, the model maintains $C$ independent estimators $F_j(\mathbf{x})$, one for each class. The evaluations of these estimators on the $n$ instances form the prediction matrix $\mathbf{F} \in \mathbb{R}^{n \times C}$, where each element is defined as $[\mathbf{F}]_{i,j} = F_j(\mathbf{x}_i)$. The residual vector $\mathbf{r}$ is transformed into a matrix $\mathbf{R} \in \mathbb{R}^{n \times C}$, calculated as $\mathbf{Y}_{\text{OneHot}} - \text{softmax}(\mathbf{F})$, where $\mathbf{Y}_{\text{OneHot}} \in \{0,1\}^{n \times C}$ is the encoded matrix and the softmax function is applied independently to each row. The remaining steps are applied identically, solving for the update coefficients that minimize the error in the vector space. The geometric angle from Eq.~(\ref{eq:theta}) is extended by vectorizing $\hat{\mathbf{P}}_\text{val} = \text{softmax}(\mathbf{F}_k(\mathbf{X}_\text{val})) \in \mathbb{R}^{n_{val} \times C}$ and $\mathbf{Y}_\text{val} \in \{0,1\}^{n_{val} \times C}$, such that
\begin{equation}
    \theta_k = \arccos\!\left(\frac{\text{vec}(\hat{\mathbf{P}}_\text{val}) \cdot \text{vec}(\mathbf{Y}_\text{val})}{\|\text{vec}(\hat{\mathbf{P}}_\text{val})\|_2 \, \|\text{vec}(\mathbf{Y}_\text{val})\|_2}\right),
\end{equation}
where $\text{vec}(\cdot)$ transforms a matrix into a column vector. The hyperparameter search is detailed in Table~\ref{tab:hyperparams_talent}. Standardization was applied to the semantic matrix in binary classification, and no transformation was used in multiclass.

\begin{table}[t]
    \centering
    \caption{Hyperparameter search space used for the evaluation in the TALENT benchmark.}
    \label{tab:hyperparams_talent}
    \resizebox{\columnwidth}{!}{
        \begin{tabular}{lcc}
          \toprule
          \textbf{Hyperparameter} & \textbf{Binary Classification} & \textbf{Multiclass Classification} \\
          \midrule
          $m$ & $8$ & $\{0.25, 0.5, 1, 2\}$ \\
          $K$ & $200$ & $200$ \\
          $\eta$ & $\{0.1, 0.5, 1\}$ & $0.1$ \\
          $d_{max}$ & $5$ & $\{1, 3\}$ \\
          $\lambda, \rho$ & $0.02$, $0$ & $0.02$, $0$ \\
          Operators & $\{+,-,\times,\text{AQ}, \sin\}$ & $\{+,-,\times,\text{AQ}, \sin\}$ \\
          $W, \gamma, P_{c}, L, \beta_{tol}, P_{d}$ & $15, 0.1, 10, 15, 0.1, 10$ & $30, 0.1, 20, 30, 0.1, 20$ \\
          \bottomrule
        \end{tabular}
    }
\end{table}

\begin{figure*}[!tp]
    \centering
    \begin{subfigure}[b]{0.47\textwidth}
        \centering
        \includegraphics[width=\linewidth]{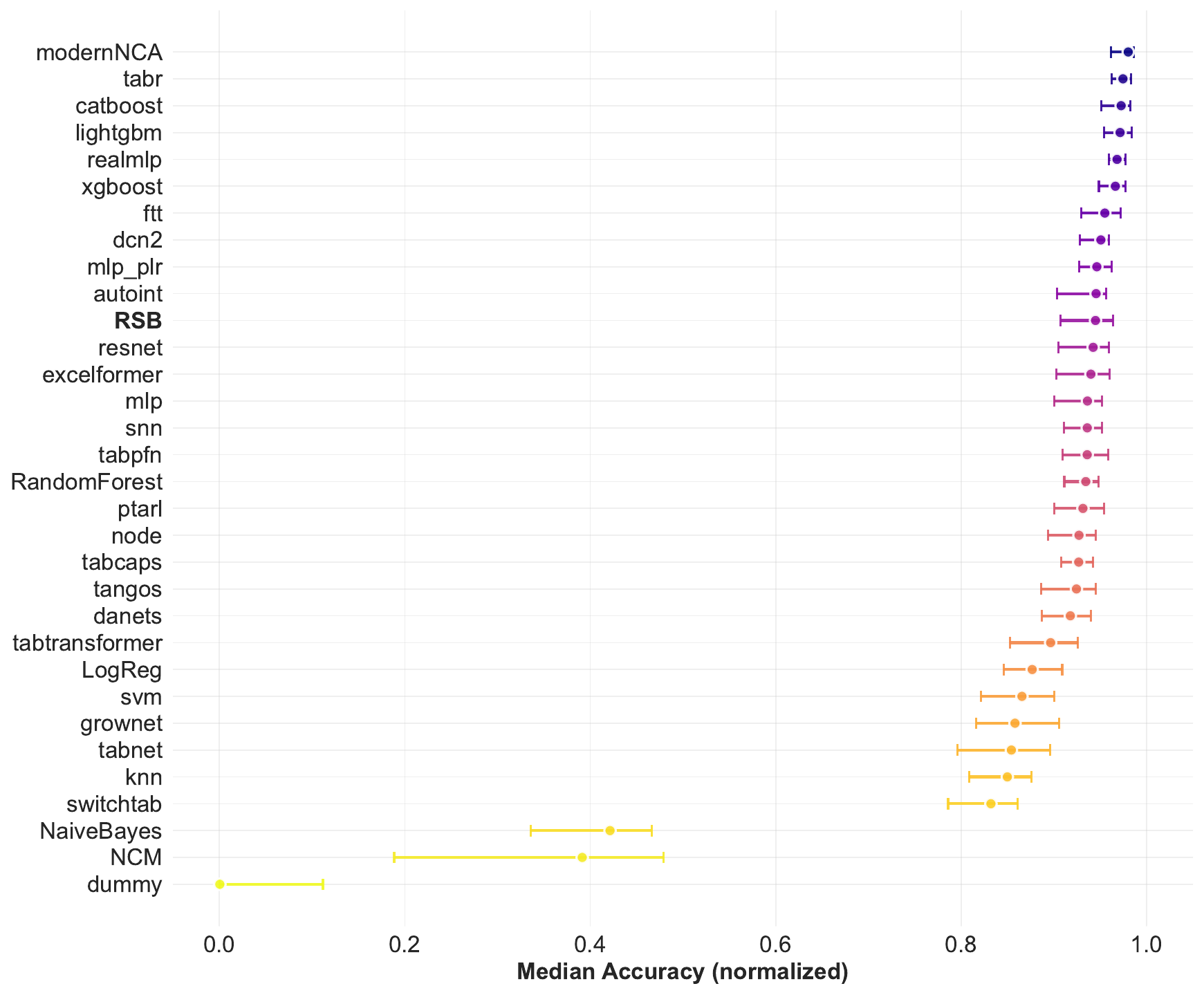}
        \caption{Binary Classification}
    \end{subfigure}
    \hfill
    \begin{subfigure}[b]{0.47\textwidth}
        \centering
        \includegraphics[width=\linewidth]{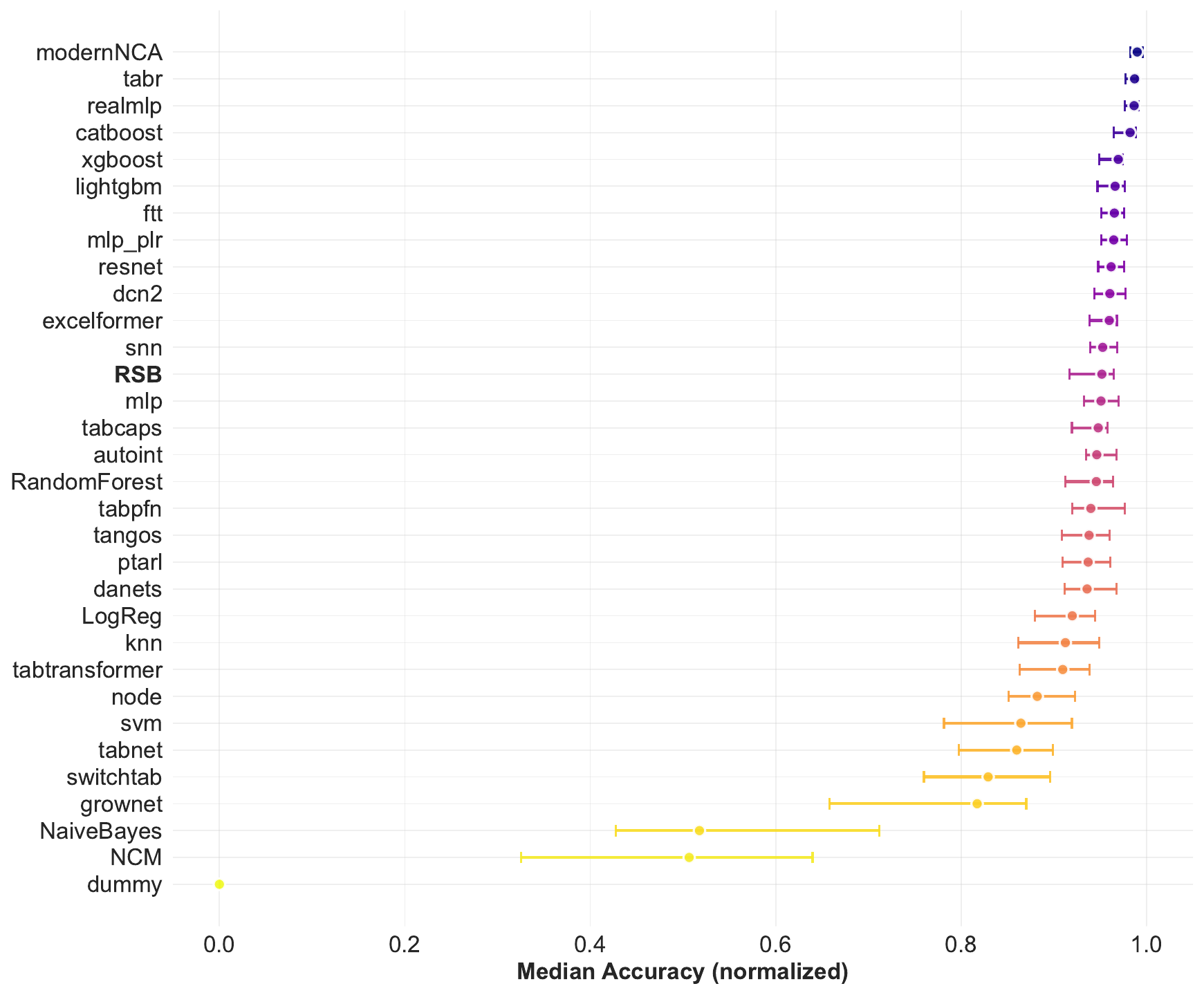}
        \caption{Multiclass Classification}
    \end{subfigure}

    \caption{Accuracy results (normalized) in the TALENT benchmark.}
    \label{fig:metrics_talent}
\end{figure*}

Figure \ref{fig:metrics_talent} illustrates the performance of the algorithms in terms of the normalized median accuracy. In binary classification, the model ranks 11\textsuperscript{th}, below XGBoost and LGBM but above ResNet, Random Forest and SVM. In multiclass classification, RSB holds the 13\textsuperscript{th} position; however, it remains competitive and comparable to XGBoost or LGBM.

\section{Discussion}\label{discussions}

In the SRBench evaluation, RSB is located at the elbow of the Pareto-optimal front between accuracy and training time; Operon is more accurate but considerably slower, whereas linear models are faster but significantly less accurate. The proposed algorithm achieved first place in average performance on the real-world datasets and the third position overall.

On the other hand, in the TALENT benchmark, RSB obtained the 11\textsuperscript{th} place in binary classification and the 13\textsuperscript{th} in multiclass classification against 31 state-of-the-art algorithms. In multiclass classification, the performance of RSB is located within a range comparable to that of methods such as XGB and LGBM. In binary classification, although it ranks below these ensembles, RSB demonstrated superior performance to architectures such as ResNet, Random Forest, and SVM, establishing it as a competitive method.

The Jaccard index presented in Figure \ref{fig:jaccard_matrix} shows similarities with FFX, EFS, RC and SBR, with notable differences: FFX expands basis functions systematically and deterministically, and EFS evolves them bottom-up before selecting them via pathwise regularized regression, whereas RSB samples them stochastically; RC creates a fixed random architecture and only optimizes the last layer, whereas RSB constructs a sequential additive model via boosting over syntax trees. Furthermore, SBR relies on semantic backpropagation and local evolution to refine trees, whereas RSB avoids evolution entirely through random generation and Elastic Net minimization. The difference with SR-Forest is greater, given that the latter employs evolutionary operators and bagging. The similarity matrix suggests that RSB does not coincide with any method in Table \ref{table:stork_classification} in all four components. Its closest neighbors, FFX, EFS and SBR (0.60), each differ from it in a single component, and are distinguished by the mechanisms discussed above.

\section{Conclusions}\label{conclusions}

This work presented RSB, a symbolic regression method that avoids generational evolution by constructing explicit additive models through boosting over populations of random syntax trees. The proposal stems from two observations: any subspace of $\mathbb{R}^n$ can be approximated given enough linearly independent vectors, and GSGP has already demonstrated that the semantics of syntax trees can be directly manipulated as vectors in that space. Combining this idea with the additive logic of ensemble methods, and with the efficiency of linear solvers, is what gives rise to RSB. Under the Stork taxonomy, the method is similar to FFX, EFS, SBR, and RC, but its sequential boosting mechanism and absence of generational evolution differentiate it from them.

From a theoretical standpoint, it is shown that, when $\lambda=0$, the boosting guarantees the descent of the training error: in regression, the norm of the residual decreases for $\eta \in (0,2)$, and in classification, the log-loss does so for $\eta \in (0,8)$. The angle-based early stopping mechanism was introduced, based on the angle between the prediction and the target. Its main advantage is that $\theta_k \in [0, \pi]$ regardless of the problem scale, which makes it applicable to regression, binary and multiclass classification without the need for additional adjustments.

Regarding performance, the experiments on SRBench showed that RSB achieves an accuracy comparable to that of advanced symbolic methods such as Operon and GP-GOMEA, but with significantly lower training times. On TALENT, the algorithm remained competitive against the leading state-of-the-art classification models, including XGBoost and LGBM.

The clearest limitation of the method is the size of the resulting models. Being an additive ensemble, the final expression can become difficult to interpret when many boosting levels are required, which distances it from the interpretability typically sought in traditional symbolic models.

As future work, it is proposed to explore post-hoc symbolic simplification strategies to reduce the complexity of the final expression without sacrificing accuracy.

\section*{Funding}
The first author was supported by a scholarship from the Secretaría de Ciencia, Humanidades, Tecnología e Innovación (SECIHTI, formerly CONAHCYT) [scholarship CVU 1105820]. The funding source had no involvement in the study design, the collection, analysis, or interpretation of data, the writing of the report, or the decision to submit the article for publication.

\section*{CRediT authorship contribution statement}
\textbf{José R. Ortega}: Conceptualization, Methodology, Software, Formal analysis, Investigation, Data curation, Visualization, Writing – original draft, Writing – review and editing.
\textbf{José Manuel Muñoz Contreras}: Investigation, Validation, Writing – review and editing.
\textbf{Leonardo Trujillo}: Conceptualization, Methodology, Supervision, Formal analysis, Writing – review and editing.

\section*{Declaration of competing interest}
The authors declare that they have no known competing financial interests or personal relationships that could have appeared to influence the work reported in this paper.

\section*{Declaration of generative AI and AI-assisted technologies in the manuscript preparation process}
During the preparation of this work the authors used Claude (Anthropic) to improve grammar and wording, and to troubleshoot LaTeX. After using this tool, the authors reviewed and edited the content as needed and take full responsibility for the content of the publication.

\section*{Data Availability Statement}
The SRBench and TALENT benchmark datasets used in this study are publicly available from their official repositories. The source code of RSB is not publicly available because it is part of ongoing doctoral research; it will be deposited in a public repository upon acceptance of the article.

\bibliographystyle{elsarticle-harv}
\bibliography{mybibfile}

\end{document}